\documentclass{article}

\usepackage[numbers]{natbib}
\usepackage[preprint]{neurips_2022}

\usepackage[dvipsnames]{xcolor}         
\definecolor{linkColor}{rgb}{0.2,0.4,0.6}
\usepackage[utf8]{inputenc} 
\usepackage[T1]{fontenc}    
\usepackage[colorlinks=true,linkcolor=linkColor,citecolor=linkColor,filecolor=linkColor,urlcolor=linkColor]{hyperref}       
\usepackage{url}            
\usepackage{booktabs}       
\usepackage{amsfonts}       
\usepackage{nicefrac}       
\usepackage{microtype}      
\usepackage{graphicx}
\usepackage{wrapfig}
\usepackage{makecell}
\usepackage{bbm}
\usepackage{float}
\usepackage{subcaption}
\usepackage{soul}
\usepackage{pifont}
\usepackage{epstopdf}
\usepackage{cuted}
\usepackage{multirow}
\usepackage{mathtools}
\usepackage{enumitem}
\usepackage{adjustbox}
\usepackage{setspace}
\usepackage{ulem}
\usepackage{capt-of}
\usepackage{algorithm}
\usepackage{algpseudocode}
\usepackage{algorithmicx}
\usepackage{wasysym}
\usepackage[most]{tcolorbox}
\usepackage{amsmath}

\definecolor{gray}{rgb}{0.5,0.5,0.5}
\definecolor{gray94}{gray}{.92}
\definecolor{gray90}{gray}{.90}
\definecolor{gray85}{gray}{.85}
\usepackage{colortbl}
\usepackage{pifont}
\usepackage{amssymb}
\usepackage{ulem} 
\usepackage{adjustbox}
\usepackage{microtype}
\usepackage{float}
\usepackage{natbib}

\definecolor{darkred}{RGB}{139,0,0}
\newcommand{\cmark}{{\color{green}\ding{51}}}%
\newcommand{\xmark}{{\color{red}\ding{55}}}%
\newcommand\ourdata{\textsc{McD}}
\newcommand\ourmodel{VisCodex}
\newcommand\ourmodelsml{VisCodex-8B}
\newcommand\ourmodellrg{VisCodex-33B}

\newcommand\ourcodeqabenchmark{InfiBench-V}

\definecolor{darkblue}{rgb}{0, 0, 0.6}
\hypersetup{
	colorlinks=true,
	linkcolor=blue,
	filecolor=blue,      
	urlcolor=blue,
}

\title{VisCodex: Unified Multimodal Code Generation via \\ Merging Vision and Coding Models}

\makeatletter
\renewcommand\@fnsymbol[1]{\dag} 
\makeatother

\author{
Lingjie Jiang$^{1,2}$~~~~Shaohan Huang$^{1}$\thanks{Corresponding Author}~~~~Xun Wu$^{1}$~~~~Yixia Li$^{1,3}$~~~~Dongdong Zhang$^{1}$~~~~Furu Wei$^{1}$ \\
$^1$ Microsoft Research ~~~
$^2$ Peking University ~~~
$^3$ Southern University of Science and Technology \\
{\href{https://aka.ms/GeneralAI}{https://aka.ms/GeneralAI}}
\vspace{-0.4cm}
\\
}

\begin{document}

\maketitle

\begin{abstract}
Multimodal large language models (MLLMs) have significantly advanced the integration of visual and textual understanding. However, their ability to generate code from multimodal inputs remains limited. In this work, we introduce VisCodex, a unified framework that seamlessly merges vision and coding language models to empower MLLMs with strong multimodal code generation abilities. Leveraging a task vector-based model merging technique, we integrate a state-of-the-art coding LLM into a strong vision-language backbone, while preserving both visual comprehension and advanced coding skills.
To support training and evaluation, we introduce the Multimodal Coding Dataset (\ourdata{}), a large-scale and diverse collection of 598k samples, including high-quality HTML code, chart image-code pairs, image-augmented StackOverflow QA, and algorithmic problems. Furthermore, we propose InfiBench-V, a novel and challenging benchmark specifically designed to assess models on visually-rich, real-world programming questions that demand a nuanced understanding of both textual and visual contexts.
%
Extensive experiments show that VisCodex achieves state-of-the-art performance among open-source MLLMs and approaches proprietary models like GPT-4o, highlighting the effectiveness of our model merging strategy and new datasets.
Our code and data are available at \url{https://github.com/JackLingjie/VisCodex}
\end{abstract}

\section{Introduction}
Multimodal large language models (MLLMs) have achieved remarkable success in recent years, demonstrating an impressive ability to understand and reason about the world by integrating information from both visual and textual domains~\citep{zhu2023minigpt, liu2023llava, Qwen-VL}. These models have pushed the boundaries of what is possible in tasks like visual question answering (VQA), image captioning, and general multimodal conversation. However, a critical and highly practical domain remains relatively underexplored: the generation of functional code from visual inputs.

This task, which we term multimodal code generation, presents a distinct set of challenges. 
It demands not only a nuanced interpretation of visual elements—such as UI layouts, data chart structures, or programming-related screenshots—but also the ability to translate these insights into syntactically flawless and functionally correct code.
While today's multimodal models excel at visual description, they often lack the deep programming knowledge required for robust code generation. This gap is critical, as many modern development tasks, like translating a UI mockup into HTML or replicating a data chart, demand a seamless fusion of visual understanding and coding proficiency.

To bridge the gap between visual perception and code generation, we introduce \ourmodel{}.
Rather than relying on costly pre-training, our approach efficiently creates a unified model by arithmetically merging the parameters of a state-of-the-art vision-language model and a dedicated coding LLM. 
Specifically, we adopt a model merging technique based on task vectors, which capture the parameter shifts resulting from fine-tuning on specific domains (e.g., vision-language, coding). By linearly combining these task vectors in the language model backbone—while keeping the vision encoder and cross-modal projection modules intact—we jointly integrate advanced code understanding and generation capabilities with nuanced visual perception. This enables the resulting model to simultaneously retain strong visual understanding and robust code generation ability, thereby significantly enhancing its performance on multimodal coding tasks.
%
Our experiments show that the merged model significantly outperforms the original vision-language model on multimodal coding tasks.

To address the lack of high-quality, large-scale training data for multimodal code generation, We introduce the \textbf{M}ultimodal \textbf{c}oding \textbf{D}ataset (\ourdata{}), a comprehensive, instruction-tuning dataset comprising 598k samples. \ourdata{} is meticulously curated from four diverse sources: (1) aesthetically enhanced and structurally sound HTML code generated from webpage screenshots, (2) high-quality chart-to-code pairs from real-world and synthetic sources, (3) image-augmented question-answer pairs from StackOverflow, and (4) foundational algorithmic coding problems to preserve core reasoning abilities.

Furthermore, to rigorously assess the real-world performance of models on multimodal coding QA task, we develop \ourcodeqabenchmark{}, a new and challenging benchmark. \ourcodeqabenchmark{} consists of visually rich programming-related questions derived from real user scenarios where the images are indispensable for arriving at the correct solution. It provides a more realistic and demanding testbed than existing benchmarks that often focus on either text-only code QA or simpler visual tasks.


Our primary contributions are threefold:
\begin{enumerate}
\item We propose \ourmodel{}, a novel approach for creating powerful multimodal code generators by merging vision and coding models, demonstrating a new and efficient path to capability enhancement.
\item We introduce \ourdata{}, a large-scale, high-quality dataset for instruction-tuning MLLMs on a wide spectrum of multimodal coding tasks, and \ourcodeqabenchmark{}, a challenging benchmark for realistic evaluation. We will release both the dataset and benchmark to facilitate reproducibility and future research.
\item We conduct extensive experiments showing that \ourmodel{} significantly outperforms existing open-source MLLMs and achieves performance competitive with leading proprietary models like GPT-4o, thereby setting a new state of the art for open-source multimodal code generation.
\end{enumerate}

\section{\ourmodel{}}
\subsection{Model Architecture}

A typical multimodal large language model (MLLM) is comprised of three primary components: a vision encoder, a language model backbone, and a projection module to connect the two modalities~\citep{li2024llava}. The vision encoder's role is to extract visual features from input images. These features are then projected by the projector module into the language embedding space. Subsequently, the language model integrates these visual representations with textual inputs, enabling multimodal understanding and reasoning.

Many existing MLLMs, such as earlier versions of Qwen-VL~\cite{Qwen-VL} and LLaVA~\cite{li2024llava}, are limited by fixed image input resolutions, which curtails their flexibility in processing images of varying sizes~\citep{Qwen-VL,liu2023llava}. To overcome this limitation, Qwen2.5-VL~\cite{Qwen2VL, Qwen2.5-VL} introduces a 2D Rotary Position Embedding (RoPE)~\citep{su2024roformer} mechanism within its Vision Transformer (ViT)~\cite{dosovitskiy2021imageworth16x16words}. This allows for the flexible processing of images with arbitrary resolutions by dynamically generating visual tokens. This approach preserves the absolute scale and spatial relationships of objects within the image. Given its enhanced flexibility and performance, we adopt the Qwen2.5-VL architecture as our foundation model.

\subsection{Model Merging}
\begin{figure*}[t]
  \centering
  \includegraphics[width=\textwidth]{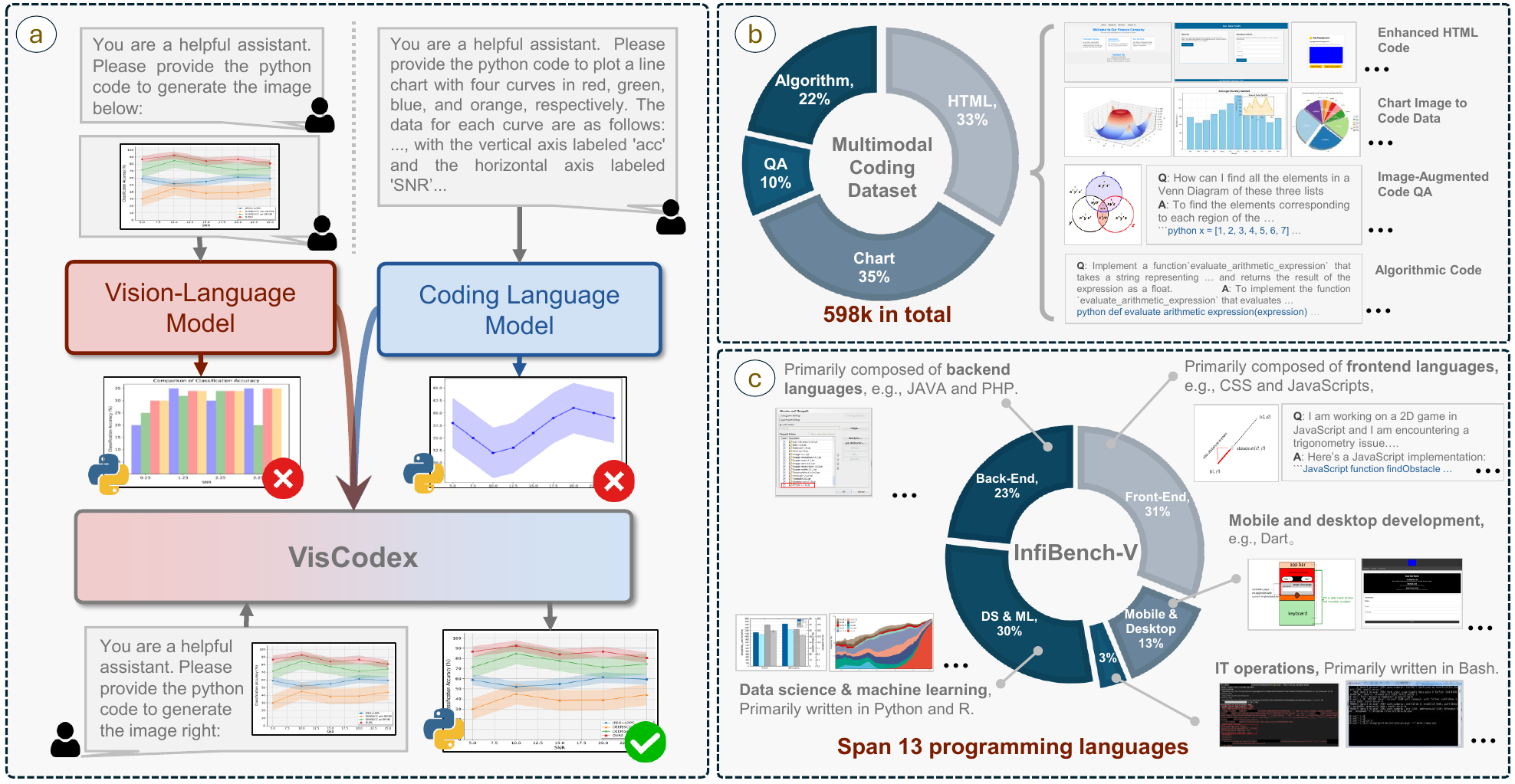}
    \caption{
        Illustration of the VisCodex pipeline.
        (a) Model merging strategy for unifying vision-language and coding LLMs;
        (b) Data distribution and representative cases of \ourdata{};
        (c) Category breakdown and representative cases of InfiBench-V.
    }
  \label{fig:main_figure}
\end{figure*}

To enhance the coding capabilities of our multimodal large language model without undertaking costly retraining from scratch, we employ model merging~\citep{jin2022dataless}. By arithmetically combining the parameters of specialized models, we can integrate distinct skills and create a unified, more versatile model without requiring access to the original training data (see Figure~\ref{fig:main_figure}~(a) for an overview of the model merging pipeline).

\textbf{Task Vectors.}
Central to model merging are \textit{task vectors}~\citep{ilharco2022editing}, which quantify parameter shifts resulting from fine-tuning a base model on a specific task. Given a pretrained base model $\theta_{\text{base}}$ and its task-specific fine-tuned variant $\theta_{\text{ft}}$, a task vector is formally defined as:

\begin{equation}
\tau_{\text{task}} = \theta_{\text{ft}} - \theta_{\text{base}}
\end{equation}

Such vectors encapsulate the parameter changes necessary for a model to specialize in a particular domain or capability and serve as modular, transferable units of knowledge across models and tasks.

\textbf{Multimodal and Code Capabilities.}
Our goal is to enhance the multimodal large language model by incorporating advanced code understanding and generation capabilities. Considering that code-related expertise predominantly resides in the language model backbone, we restrict our merging process to this component. We retain the original visual encoder and cross-modal projection modules unchanged to preserve the intrinsic visual understanding capabilities of the MLLM.

Specifically, we define the task vector for the language model component of the Vision-Language Model (VLM) as:

\begin{equation}
\tau_{\text{vlm}} = \theta_{\text{vlm}} - \theta_{\text{base}}
\end{equation}

where $\tau_{\text{vlm}}$ captures the parameter shift that enables the language model to effectively handle multimodal inputs by jointly processing visual and textual information.

Analogously, we define the task vector for the coding model, encapsulating its capability for code comprehension and generation:

\begin{equation}
\tau_{\text{code}} = \theta_{\text{code}} - \theta_{\text{base}}
\end{equation}

\textbf{Merging Strategy.}
Following the linear merging method of~\cite{ilharco2022editing} and its application to enhancing multimodal mathematical reasoning in~\cite{chen2025bring}, we adopt a similar strategy to transfer code reasoning abilities into an MLLM. The updated language model parameters, combining both multimodal and code-related knowledge, are computed as follows:

\begin{equation}
\label{eq:merging_formula}
\theta_{\text{\ourmodel{}}} = \theta_{\text{base}} + \lambda \tau_{\text{vlm}} + (1 - \lambda) \tau_{\text{code}}
\end{equation}

where the hyperparameter $\lambda \in [0, 1]$ controls the trade-off between retaining original multimodal representations and integrating new code expertise. $\theta_{\text{\ourmodel{}}}$ is the initialization of the parameters of our \ourmodel{}.
%
%
%

\textbf{Implementation Details.}
Our model merging process targets only the language backbone of the VLM, leaving the vision encoder and cross-modal projection modules unaltered. This selective merging approach allows for a clear attribution of performance gains while significantly reducing computational overhead. To construct our primary code task vector ($\tau_{code}$), we select a coding model that shares the same architectural foundation as the VLM's language backbone. Since Qwen2.5-VL's language model is derived from Qwen2.5, we utilize OpenCodeReasoning-Nemotron-1.1-7B~\citep{ahmad2025opencodereasoning}. For our 33B scale model, we correspondingly use the OpenCodeReasoning-Nemotron-1.1-32B~\citep{ahmad2025opencodereasoning} variant.
Furthermore, in our ablation studies, we create and evaluate code task vectors from two other prominent code-specialized models, Qwen2.5-Coder-7B-Instruct~\citep{hui2024qwen2} and OpenThinker2-7B~\citep{guha2025openthoughtsdatarecipesreasoning}, to verify the effectiveness of merging with a code-specialized task vector, as shown in Table~\ref{tab:ablation_merge_coder_revised}
\subsection{Model Training}
After model merging, we perform supervised fine-tuning on our Multimodal Coding Dataset (\ourdata{}), further aligning the merged model with multimodal coding tasks. To efficiently leverage both the pretrained visual grounding and newly integrated code abilities, we freeze the vision encoder and projection modules, fine-tuning only the language model backbone.

\section{\ourdata{} Dataset}
We introduce the Multimodal Coding Dataset (\ourdata), a new large-scale dataset designed for instruction-tuning multimodal models on coding tasks. \ourdata{} is constructed from four primary components, each targeting a distinct aspect of multimodal code understanding and generation:

\begin{enumerate}
    \item \textbf{Enhanced HTML Code}: We generate aesthetically and structurally improved HTML code by redesigning and augmenting existing webpages.
    \item \textbf{Chart Image-Code Pairs}: We construct high-quality chart-code pairs by sourcing Python matplotlib code from GitHub, which are then refined through a multi-stage filtering and rewriting process.
    \item \textbf{Image-Augmented Code QA}: We extract real-world, image-augmented question and answer pairs from StackOverflow and subject them to rigorous cleaning and refinement.
    \item \textbf{Algorithmic Code}: We aggregate and curate data from established algorithmic coding datasets to preserve and enhance the model's core reasoning and problem-solving abilities.
\end{enumerate}
%
%
The data distribution and representative cases for the four domains are illustrated in Figure~\ref{fig:main_figure}~(b), and more comprehensive statistics are available in the Appendix~\ref{app:mcd_data_statistics}. The following sections describe each component in detail.

\subsection{Enhanced HTML Code}

A review of the existing Web2Code dataset~\citep{web2code2024} revealed several shortcomings, including broken image links, rudimentary CSS, and visually unappealing designs. Our initial approach to address these issues involved using GPT-4o to directly rewrite the existing HTML code. However, this method proved suboptimal, as the constraints of the original code structure frequently led to rendering artifacts and visually incongruous layouts.

To overcome this, we adopted a novel, image-driven generation pipeline. We first curated 560,000 webpage images from Web2Code to serve as stylistic seeds. GPT-4o was then prompted to design entirely new webpages inspired by these seeds. The resulting HTML was rendered using Playwright\footnote{https://github.com/microsoft/playwright-python} to capture screenshots. A rigorous filtering pipeline was then applied to discard rendering failures, images with anomalous dimensions, and other visual artifacts. This process yielded 200,000 high-quality, newly generated code-image pairs. These pairs were subsequently converted into an instruction-following format using the framework provided by Web2Code.


\subsection{Chart Image to Code Data}

To build a diverse and high-quality chart dataset, we incorporate both synthetic and real-world data sources. For synthetic data, we include the 164,000 synthetic Chart2Code samples released by ChartCoder~\citep{zhao2025chartcoderadvancingmultimodallarge} as part of our training data. For real-world data, we curated 46,000 chart-code pairs from GitHub.

Inspired by data engineering strategies~\citep{ding2023enhancing, chiang2023vicuna, xu2023wizardlm}, we first collected 129,000 real-world Python matplotlib scripts from GitHub. 
This raw data, however, suffered from significant quality issues, including non-executable code, inconsistent formatting, and potentially harmful snippets. To mitigate these issues, we employed GPT-4o to systematically rewrite and normalize the code, while simultaneously classifying each script by chart type. Subsequently, a multi-stage, rule-based filtering pipeline was applied to eliminate low-quality samples. This pipeline removed scripts that failed to execute, produced blank or improperly sized images, or generated visually corrupted outputs (e.g., heavily pixelated charts). As a final quality assurance step, we leveraged GPT-4o to score the aesthetic and functional quality of the generated charts, retaining the top 46,000 high-quality image-code pairs.

The final dataset combines the 164,000 synthetic samples with our 46,000 curated real-world examples, resulting in a comprehensive collection of 210,000 chart image-code pairs for instruction tuning.



\subsection{Image-Augmented Code QA}

StackOverflow represents a rich repository of real-world, code-centric QA data, particularly valuable when augmented with illustrative images. Our collection process involved crawling StackOverflow for QA threads containing images, followed by an initial filtering step to retain only those with an accepted answer containing either Python or HTML code.

A rigorous data cleaning pipeline was implemented to ensure quality, removing entries with excessively short or verbose answers, invalid URLs, broken image links, and blank or oversized images. We also identified that many accepted answers were suboptimal for training, being either too terse for clarity or overly verbose. To address this, we utilized GPT-4o to refine these answers by removing sensitive content, rewriting unclear sections, and enhancing overall conciseness and clarity. This multi-stage pipeline yielded a final dataset of 59,000 high-quality, image-augmented StackOverflow QA pairs suitable for instruction tuning.




\subsection{Algorithmic Code}

To maintain the model’s proficiency in algorithmic reasoning and code generation, we incorporate algorithm-related code data from Kodcode~\citep{xu2025kodcode}. Specifically, we select samples from five categories: LeetCode~\citep{hartford2023leetcode}, Codeforces~\citep{jur1cek2023codeforces}, TACO~\citep{li2023taco}, Code Contests~\citep{li2022competition}, and Algorithm~\citep{thealgorithms2023python, keon2017pythonic}. The final collection contains 129,000 algorithm-related instruction-following examples.

\section{\ourcodeqabenchmark{}}

We introduce \ourcodeqabenchmark{}, a new benchmark designed to evaluate the ability of multimodal large language models to answer complex programming questions that integrate both text and images. While existing benchmarks like InfiBench~\citep{li2024infibench} focus on text-based code QA, \ourcodeqabenchmark{} is specifically constructed to assess multimodal reasoning, where visual context is critical to formulating a correct answer.


\subsection{Data Curation}

Our benchmark is built upon a rigorous, multi-stage curation pipeline using data from Stack Overflow. The process began by scraping an initial set of approximately 1 million image-based questions that included a community-verified ``accepted answer'' to ensure solution quality. We then narrowed this pool to 40,000 recent and high-engagement questions. The most critical refinement step involved using GPT-4o to isolate samples where the image is indispensable, filtering out questions solvable by text alone. This yielded a core set of 10,000 high-relevance, multimodal questions.

We categorized these samples based on programming domain and, guided by the class distribution and sampling principles of InfiBench, domain experts manually selected 322 questions to form the final benchmark. These span 13 programming languages, each mapped to one of five high-level categories: front-end, back-end, data science \& machine learning (DS\&ML), mobile and desktop development, and IT operations (ITOps). 
The detailed category breakdown and representative cases are shown in Figure~\ref{fig:main_figure}~(c)

To ensure quality and prevent model memorization in pre-training, we implemented a prompt paraphrasing process. Domain experts rewrote each question in a concise and directive manner while preserving its semantic content.
Each question is also annotated with its evaluation category and associated metrics, including a set of key phrases and a reference answer to support robust scoring.

\subsection{Evaluation Criteria}

To objectively assess the quality of answers across a diverse range of question types, we adopt a three-pronged evaluation strategy inspired by InfiBench.

\textbf{Keyword Matching.} 
We observed that for a majority of questions, answer quality is closely tied to the presence of specific keywords. Our domain experts craft a set of rules for each question, specifying essential terms and phrases. To capture nuanced requirements, these rules can be simple checks, regular expressions, or complex logical statements. When multiple keywords are required, they can be individually weighted to ensure that the most critical components of the answer contribute more significantly to the final score.


\textbf{Unit Testing}
For questions where the answer is primarily a block of code, we verify its correctness using unit tests. To facilitate automated evaluation, domain experts supplement the question with precise requirements, like function names and expected I/O formats. They also provide the necessary setup and teardown scripts, creating a complete and executable environment for programmatic validation.

\textbf{GPT-4o Judge.}
For questions that rely heavily on natural language understanding, we leverage GPT-4o to score MLLM responses by comparing them with the accepted reference answer. The evaluation considers both answer correctness and completeness across two dedicated scoring dimensions.

\section{Experimental Setup}

\noindent\textbf{Evaluated Benchmarks.}  We evaluate our model on four multimodal benchmarks to assess a range of multimodal-related coding skills:

\begin{itemize}
    \item \textbf{Design2Code~\citep{si2024design2code}}: This benchmark measures the ability to translate visual UI designs into executable code. We report the average performance on both Low-Level (Low-L) features (Block, Text, Position, Color) and High-Level (High-L) semantic fidelity.
    \item  \textbf{ChartMimic~\citep{shi2024chartmimic}}: This benchmark evaluates the generation of chart specifications from images. We adopt the Direct Mimic task on the test-mini subset and report both Low-Level (Low-L) and GPT-4o-assessed High-Level (High-L)  scores.
    \item \textbf{MMCode~\citep{li2024mmcode}}: This benchmark assesses algorithmic problem-solving in visually rich contexts. Performance is measured by pass@1 accuracy~\citep{chen2021evaluating}.
    \item \textbf{\ourcodeqabenchmark{} (Ours)}: For our proposed benchmark, we report the average score across all defined evaluation metrics.
\end{itemize}

\noindent\textbf{Training Settings.} 


In our main experiments with the 8B model, which uses the code task vector from OpenCodeReasoning-Nemotron-1.1-7B, we determined the optimal merge coefficient $\lambda$ by evaluating performance on the MMCode benchmark. From a set of candidate values $\{0.7, 0.8, 0.85, 0.9\}$, we selected $\lambda=0.7$. According to our merging formula~\ref{eq:merging_formula}, this applies a weight of 0.7 to the vision-language task vector ($\tau_{vlm}$) and 0.3 to the code task vector ($\tau_{code}$). For the 33B variant, we directly set $\lambda=0.85$ without additional tuning due to resource constraints. 
Detailed training hyperparameters and training costs are provided in the Appendix~\ref{app:hyperparameters}.
%


\section{Experimental Results}

\subsection{Main Results}

\begin{table*}[h]
\centering
\renewcommand\tabcolsep{6pt}
\resizebox{\linewidth}{!}{
\begin{tabular}{l c cc cc c c c}
\toprule
\textbf{Model} & \textbf{Size}
& \multicolumn{2}{c}{\textbf{Design2Code}}
& \multicolumn{2}{c}{\textbf{ChartMimic}}
& \textbf{MMCode}
& \textbf{InfiBench-V}
& \textbf{Average} \\
\cmidrule(lr){3-4} \cmidrule(lr){5-6} \cmidrule(lr){7-7} \cmidrule(lr){8-9}
& & Low-L & High-L & Low-L & High-L & pass@1 & Acc. & \\
\midrule
\multicolumn{9}{l}{\textit{Proprietary Models}} \\
\midrule
GPT-4o-mini & - & 85.8 & 87.3 & 68.4 & 68.5 & 12.2 & 71.9 & 65.7 \\
GPT-4o & - & 90.2 & 90.4 & 79.0 & 83.5 & 17.0 & 79.9 & 73.3 \\
\midrule
\multicolumn{9}{l}{\textit{Open-Source Small Language Models}} \\
\midrule
MiniCPM-V-2\_6 & 8B & 78.1 & 84.2 & 21.8 & 45.2 & 3.8 & 45.3 & 46.4 \\
InternVL3-8B & 8B & 85.3 & 87.6 & 43.1 & 47.2 & 6.8 & 66.1 & 56.0 \\
Qwen2.5-VL-7B-Instruct & 8B & 83.4 & 87.6 & 39.5 & 38.3 & 5.3 & 54.0 & 51.4 \\
Llama-3.2-11B-Vision-Instruct & 11B & 72.7 & 84.8 & 27.7 & 26.5 & 2.3 & 52.7 & 44.4 \\
InternVL3-14B & 15B & 82.9 & 88.3 & 53.9 & 55.0 & \textbf{11.4} & 70.5 & 60.3 \\
\textbf{\ourmodel-8B} & \textbf{8B} & \textbf{90.1} & \textbf{90.9} & \textbf{74.8} & \textbf{74.1} & 11.0 & \textbf{72.1} & \textbf{68.8} \\
\midrule
\multicolumn{9}{l}{\textit{Open-Source Large Language Models}} \\
\midrule
Qwen2.5-VL-32B-Instruct & 33B & 88.0 & 89.4 & 72.5 & 68.7 & 13.7 & 73.0 & 67.6 \\
llava-onevision-qwen2-72b & 73B & 75.2 & 85.7 & 55.8 & 52.1 & 5.7 & 64.7 & 56.5 \\
Qwen2.5-VL-72B-Instruct & 73B & 86.9 & 88.7 & 66.7 & 68.7 & 15.2 & 75.2 & 66.9 \\
InternVL3-78B & 78B & 85.3 & 89.1 & 64.9 & 64.2 & 14.4 & 77.3 & 65.9 \\
\textbf{\ourmodel-33B} & \textbf{33B} & \textbf{90.5} & \textbf{91.1} & \textbf{79.3} & \textbf{78.5} & \textbf{15.6} & \textbf{78.6} & \textbf{72.3} \\
\bottomrule
\end{tabular}}
\caption{Performance comparison between proprietary and open-source models across various benchmarks.  Low-L stands for Low-Level features (e.g., Block, Text, Position), and High-L stands for High-Level semantic fidelity. Best results are in \textbf{bold}.}
\label{tbl:final_benchmark_full}
\end{table*}
As shown in Table \ref{tbl:final_benchmark_full}, our models achieve state-of-the-art performance across all evaluated multimodal coding benchmarks. Our smaller model, \ourmodelsml{}, not only outperforms all open-source models in its size class (7-15B) but also surpasses the proprietary GPT-4o-mini, with an average score of 68.8. Our larger model, \ourmodellrg{}, further establishes its superiority by achieving an average score of 72.3, which is on par with the state-of-the-art proprietary model, GPT-4o (73.3). These results demonstrate that our \ourmodel{} family sets a new standard for open-source multimodal code generation. 

Our models show exceptional strength in UI and chart understanding. On the Design2Code benchmark, both \ourmodelsml{} (90.1/90.9) and \ourmodellrg{} (90.5/91.1) achieve scores comparable to or exceeding GPT-4o. On ChartMimic, our models also secure the top positions among open-source models, demonstrating robust visual data translation capabilities.

\subsection{Analysis}
\label{sec:ablation}


\begin{table}[h]
\centering
\resizebox{0.7\linewidth}{!}{%
\begin{tabular}{l|cc|cc|c}
\toprule
\textbf{Method} & \multicolumn{2}{c|}{\textbf{Design2Code}} & \multicolumn{2}{c|}{\textbf{ChartMimic}} & \textbf{MMCode} \\
& Low-L & High-L & Low-L & High-L & pass@1 \\
\midrule
\textbf{\ourmodelsml{}} & \textbf{90.1} & \textbf{90.9} & \textbf{74.8} & \textbf{74.1} & \textbf{11.0} \\
w/o model merge & 89.6 & 90.7 & 73.4 & 70.6 & 6.8 \\
\midrule
\textbf{\ourmodellrg{}} & \textbf{90.5} & \textbf{91.1} & \textbf{79.3} & \textbf{78.5} & \textbf{15.6} \\
w/o model merge & 89.7 & 90.7 & 78.4 & 77.4 & 14.4 \\
\bottomrule
\end{tabular}%
}
\caption{Ablation on model merging for VisCodex. Results compare models with and without the merging strategy.}
\label{tab:ablation_merge_coder_wo}
\end{table}

\noindent\textbf{Efficacy of the Model Merging.}
As demonstrated in Table~\ref{tab:ablation_merge_coder_wo}, model merging yields consistent performance gains across all benchmarks and scales. At the 8B scale, merging improves Design2Code (90.1 vs. 89.6), ChartMimic (74.8 vs. 73.4), and MMCode (11.0 vs. 6.8). The 33B model shows similar enhancements. The most significant improvements on ChartMimic and MMCode confirm that this strategy effectively augments code-generation capabilities while preserving visual understanding.


\begin{table}[h]
\centering
\resizebox{0.7\linewidth}{!}{%
\begin{tabular}{l|cc|cc|c}
\toprule
\textbf{Backbone LLM} & \multicolumn{2}{c|}{\textbf{Design2Code}} & \multicolumn{2}{c|}{\textbf{ChartMimic}} & \textbf{MMCode} \\
& Low-L & High-L & Low-L & High-L & pass@1 \\
\midrule
Baseline (Qwen2.5-VL) & 83.4 & 87.6 & 39.5 & 38.3 & 5.3 \\
\midrule
\multicolumn{6}{l}{\textit{General LLM}} \\
\midrule
Qwen2.5-7B-Instruct & 89.5 & 90.7 & 73.2 & 72.5 & 6.8 \\
\midrule
\multicolumn{6}{l}{\textit{Code LLM}} \\
\midrule
OpenThinker2-7B & 90.2 & 91.0 & 74.3 & 73.8 & 8.0 \\
Qwen2.5-Coder-7B & 90.0 & 90.7 & 75.1 & 74.5 & 8.4 \\
Nemotron-1.1-7B & 90.1 & 90.9 & 74.8 & 74.1 & 11.0 \\
\bottomrule
\end{tabular}%
}
\caption{Ablation on Backbone LLM choice. Performance when merging the same multimodal backbone with either a general-purpose or code-specialized LLM.}
\label{tab:ablation_merge_coder_revised}
\vspace{-1em}
\end{table}

\paragraph{Effect of Different Code LLMs in Merge.}

As shown in the Table~\ref{tab:ablation_merge_coder_revised}, we study how the choice of the merged LLM affects performance. All code-pretrained LLMs present consistent gains across all benchmarks compared to general-purpose LLM.
Compared to the general LLM, OpenThinker2-7B and Qwen2.5-Coder-7B improve both Design2Code and ChartMimic, while  Nemotron-1.1-7B further boosts MMCode pass@1 from 6.8 to 11.0. 
These results indicate that code-oriented pretraining is crucial for robust multimodal code generation, enhancing executable correctness while maintaining strong visual grounding and UI-to-code translation.

\begin{table}[h]
\centering
\resizebox{0.7\linewidth}{!}{%
\begin{tabular}{l|cc|cc|c}
\toprule
\textbf{Strategy} & \multicolumn{2}{c|}{\textbf{Design2Code}} & \multicolumn{2}{c|}{\textbf{ChartMimic}} & \textbf{MMCode} \\
& Low-L & High-L & Low-L & High-L & pass@1 \\
\midrule
Baseline & 83.4 & 87.6 & 39.5 & 38.3 & 5.3 \\
\midrule
Replace (1-stage) & 88.7 & 90.7 & 70.4 & 69.2 & 11.0 \\
Replace (2-stage) & 88.2 & 90.6 & 73.4 & 70.9 & 11.0 \\
\textbf{Model Merge (Ours)} & \textbf{90.1} & \textbf{90.9} & \textbf{74.8} & \textbf{74.1} & \textbf{11.0} \\
\bottomrule
\end{tabular}%
}

\caption{Performance comparison of model merging vs. backbone replacement. The "Replace (1-stage)" strategy directly replaces the LLM backbone in a single stage. The "Replace (2-stage)" strategy first trains a projector, then fine-tunes the full MLLM (ViT, projector, and LLM).}
\label{tab:ablation_merge_vs_replace_revised}
\vspace{-1em}
\end{table}

\noindent\textbf{Effectiveness of the Model Merge Strategy.}
To evaluate the effectiveness of our proposed model merge strategy compared to direct backbone replacement, we conducted comparative experiments using two distinct approaches: (i) directly replacing the LLM backbone of Qwen2.5-VL-7B-Instruct with OpenCodeReasoning-Nemotron-1.1-7B~\citep{ahmad2025opencodereasoning}, and (ii) employing the two-stage training procedure from LLaVA-OneVision~\citep{li2024llava}, which initially trains the projector on BLIP-558K, followed by joint fine-tuning of the ViT, projector, and LLM on \ourdata{}.

Our results indicate that the model merge strategy achieves overall superior performance across the evaluated tasks, as shown in Table~\ref{tab:ablation_merge_vs_replace_revised}. It demonstrates particularly strong gains on visually-intensive benchmarks such as Design2Code and ChartMimic, where successful code generation heavily relies on accurate visual-semantic alignment. This is because directly replacing the LLM backbone often disrupts previously learned visual grounding. In contrast, the model merge approach preserves these visual alignment abilities while simultaneously incorporating enhanced code generation capabilities. This confirms the effectiveness of model merging in maintaining multimodal comprehension and boosting performance in multimodal coding tasks.


\begin{table}[h]
    \centering
    \resizebox{0.7\linewidth}{!}{
    \renewcommand\tabcolsep{6.0pt}
    \begin{tabular}{lcccc|c}
    \toprule
        \textbf{Data} & \textbf{Block-Match} & \textbf{Text} & \textbf{Position} & \textbf{Color} & \textbf{CLIP} \\ 
        \midrule
        Baseline & 85.4 & 95.8 & 77.3 & 75.3 & 87.6 \\
        WebCode2M & 82.1 & 96.0 & 72.4 & 72.6 & 86.3 \\
        Web2Code & 84.4 & 93.4 & 76.2 & 79.6 & 88.8 \\
        \ourdata{} & \textbf{89.6} & \textbf{97.2} & \textbf{84.7} & \textbf{86.8} & \textbf{90.7} \\
    \bottomrule
    \end{tabular}
    }
    \caption{Performance comparison of \ourdata{} with WebCode2M and Web2Code on the Design2Code benchmark.}
    \label{tab:web2code_comp}
\vspace{-1em}
\end{table}



\noindent\textbf{Comparison with Existing Web2Code Datasets.} As demonstrated in Table~\ref{tab:web2code_comp}, our dataset outperforms prior Web2Code corpora across all metrics. Compared to Webcode2M~\cite{gui2025webcode2m}, and Web2Code~\cite{web2code2024}, \ourdata{} achieves the highest scores on both low-level (Block-Match, Text, Position, and Color), and high-level evaluation. The gains are especially notable on layout-sensitive metrics  and visual fidelity, indicating that \ourdata{} provides more accurate structural alignment and visual grounding for UI-to-code generation, while also improving semantic consistency.
%
We further evaluate the generalizability of \ourdata{} in Appendix~\ref{app:exp_generalize}, where we present supervised fine-tuning results on two models with different architectures, InternVL3-8B and llava-llama3.1-8b

   v~\citep{zhang2024mmllmsrecentadvancesmultimodal}, together with additional experiments demonstrating that our code model merging strategy is effective beyond the Qwen2.5 backbone.

\begin{figure}[h]
\centering

\begin{minipage}{0.185\textwidth}
  \centering
  \fbox{\includegraphics[width=\linewidth, height=0.065\textheight]{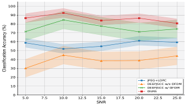}}
\end{minipage}\hspace{0.013\textwidth}
\begin{minipage}{0.185\textwidth}
  \centering
  \fbox{\includegraphics[width=\linewidth, height=0.065\textheight]{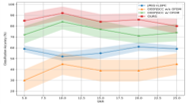}}
\end{minipage}\hspace{0.013\textwidth}
\begin{minipage}{0.185\textwidth}
  \centering
  \fbox{\includegraphics[width=\linewidth, height=0.065\textheight]{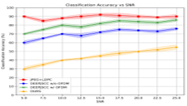}}
\end{minipage}\hspace{0.013\textwidth}
\begin{minipage}{0.185\textwidth}
  \centering
  \fbox{\includegraphics[width=\linewidth, height=0.065\textheight]{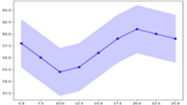}}
\end{minipage}\hspace{0.013\textwidth}
\begin{minipage}{0.185\textwidth}
  \centering
  \fbox{\includegraphics[width=\linewidth, height=0.065\textheight]{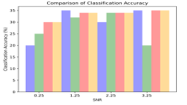}}
\end{minipage}

\vspace{0.08em}

\begin{minipage}{0.185\textwidth}
  \centering
  \fbox{\includegraphics[width=\linewidth, height=0.055\textheight]{cases/g2.png}}
\end{minipage}\hspace{0.013\textwidth}
\begin{minipage}{0.185\textwidth}
  \centering
  \fbox{\includegraphics[width=\linewidth, height=0.055\textheight]{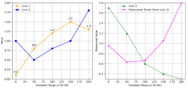}}
\end{minipage}\hspace{0.013\textwidth}
\begin{minipage}{0.185\textwidth}
  \centering
  \fbox{\includegraphics[width=\linewidth, height=0.055\textheight]{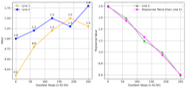}}
\end{minipage}\hspace{0.013\textwidth}
\begin{minipage}{0.185\textwidth}
  \centering
  \fbox{\includegraphics[width=\linewidth, height=0.055\textheight]{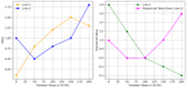}}
\end{minipage}\hspace{0.013\textwidth}
\begin{minipage}{0.185\textwidth}
  \centering
  \fbox{\includegraphics[width=\linewidth, height=0.055\textheight]{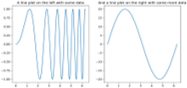}}
\end{minipage}

\vspace{0.08em}

\begin{minipage}{0.185\textwidth}
  \centering
  \fbox{\includegraphics[width=\linewidth, height=0.07\textheight]{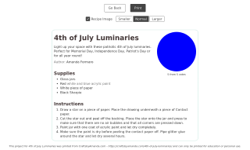}}
\end{minipage}\hspace{0.013\textwidth}
\begin{minipage}{0.185\textwidth}
  \centering
  \fbox{\includegraphics[width=\linewidth, height=0.07\textheight]{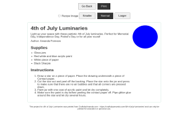}}
\end{minipage}\hspace{0.013\textwidth}
\begin{minipage}{0.185\textwidth}
  \centering
  \fbox{\includegraphics[width=\linewidth, height=0.07\textheight]{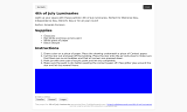}}
\end{minipage}\hspace{0.013\textwidth}
\begin{minipage}{0.185\textwidth}
  \centering
  \fbox{\includegraphics[width=\linewidth, height=0.07\textheight]{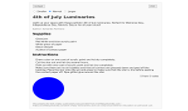}}
\end{minipage}\hspace{0.013\textwidth}
\begin{minipage}{0.185\textwidth}
  \centering
  \fbox{\includegraphics[width=\linewidth, height=0.07\textheight]{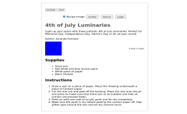}}
\end{minipage}

\vspace{0.08em}

\begin{minipage}{0.185\textwidth}
  \centering
  \fbox{\includegraphics[width=\linewidth, height=0.085\textheight]{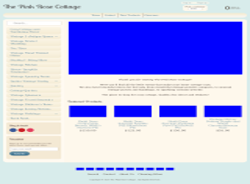}}
\end{minipage}\hspace{0.013\textwidth}
\begin{minipage}{0.185\textwidth}
  \centering
  \fbox{\includegraphics[width=\linewidth, height=0.085\textheight]{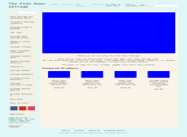}}
\end{minipage}\hspace{0.013\textwidth}
\begin{minipage}{0.185\textwidth}
  \centering
  \fbox{\includegraphics[width=\linewidth, height=0.085\textheight]{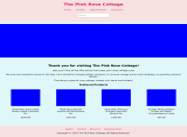}}
\end{minipage}\hspace{0.013\textwidth}
\begin{minipage}{0.185\textwidth}
  \centering
  \fbox{\includegraphics[width=\linewidth, height=0.085\textheight]{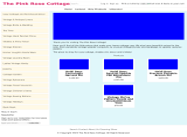}}
\end{minipage}\hspace{0.013\textwidth}
\begin{minipage}{0.185\textwidth}
  \centering
  \fbox{\includegraphics[width=\linewidth, height=0.085\textheight]{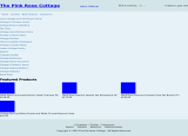}}
\end{minipage}

\vspace{0.5em}

\begin{minipage}{0.185\textwidth}
  \centering
  Ground Truth
\end{minipage}\hspace{0.012\textwidth}
\begin{minipage}{0.185\textwidth}
  \centering
  \textcolor{darkred}{\bf VisCodex-8B (Ours)}
\end{minipage}\hspace{0.012\textwidth}
\begin{minipage}{0.185\textwidth}
  \centering
  GPT-4o-mini
\end{minipage}\hspace{0.012\textwidth}
\begin{minipage}{0.185\textwidth}
  \centering
  InternVL3-78B
\end{minipage}\hspace{0.012\textwidth}
\begin{minipage}{0.185\textwidth}
  \centering
  Qwen2.5-VL-7B
\end{minipage}

\vspace{0.5em}

\caption{Case study comparing VisCodex-8B and baseline models on Design2Code and ChartMimic tasks, demonstrating the superior multimodal code generation capabilities of VisCodex-8B.}
\label{fig:case_study}
\end{figure}

\subsection{Case Study}

We further conducted case studies to qualitatively compare the performance of VisCodex-8B against GPT-4o-mini, InternVL3-78B, and Qwen2.5-VL-7B on the ChartMimic and Design2Code benchmarks. As shown in Figure~\ref{fig:case_study}, VisCodex-8B consistently generates outputs that more closely match the ground truth in both chart reconstruction and HTML generation tasks, surpassing the fidelity of results produced by GPT-4o-mini and other open-source baselines. These observations underscore VisCodex-8B's superior multimodal code generation capabilities. For additional case studies on MMCode, InfiBench-V, and further examples, please refer to Appendix~\ref{app:additional_cases}.

\section{Related Work}

\subsection{Multimodal Large Language Models}
Recent advancements in Multimodal Large Language Models (MLLMs)~\citep{awadalla2023openflamingo, cao2024dualfocus,chen2023pali,dai2023instructblip,driess2023palm,huang2024language,li2023mimic,zhu2023minigpt,peng2023kosmos,pan2023kosmos}  have been achieved by integrating large language models (LLMs) with vision components, enabling rich text-image interactions. For example, MiniGPT-4~\citep{zhu2023minigpt} and InstructBLIP~\citep{chen2023pali} extend language-only instruction tuning to multimodal tasks through the use of Q-Former~\citep{li2023blip}. Meanwhile, models such as LLaVA~\citep{liu2023llava,liu2024llavanext,li2024llava}, Qwen2.5-VL~\citep{Qwen-VL,Qwen2VL, Qwen2.5-VL}, and InternVL3~\citep{zhu2025internvl3exploringadvancedtraining} use MLPs to map visual features into the LLM embedding space while simultaneously fine-tuning the model with LLM objectives. These innovations have led to remarkable success in tasks such as VQA~\citep{antol2015vqa} and chart understanding~\citep{han2023chartllama,zeng2024advancing}, demonstrating the power of MLLMs in handling complex multimodal reasoning tasks.

\subsection{Multimodal Code Generation}
The ability of MLLMs to generate code has attracted increasing attention in recent years. %
Design2Code~\citep{si2024design2code} evaluates the HTML generation capabilities of MLLMs. Extending earlier datasets like WebSight~\citep{laurenccon2024unlocking} and Pix2Code~\citep{beltramelli2018pix2code}, Web2Code~\citep{web2code2024}, Webcode2M~\cite{gui2025webcode2m} provides a webpage-to-code dataset to improve HTML generation. Benchmarks like MMCode~\citep{li2024mmcode} and Human-V~\citep{zhang2024humaneval} focus on assessing MLLMs in algorithmic coding tasks that incorporate visual inputs. Similarly, ChartMimic~\citep{shi2024chartmimic} and Plot2Code~\citep{wu2024plot2code} evaluate MLLMs' capabilities to translate raw data into scientific charts. ChartCoder~\citep{zhao2025chartcoder} addresses chart generation explicitly through a large dataset of 160k examples. Additionally, CodeV~\citep{zhang2024codev} integrates visual data to improve large language models' problem-solving abilities.
Despite recent progress, to the best of our knowledge, existing work falls short of providing a complete and unified solution to multimodal code generation.


\subsection{Model Merging for MLLMs}
Model merging has become a widely used approach for integrating the capabilities of multiple models within the parameter space. A basic method involves simple weighted averaging~\citep{wortsman2022model}, while more advanced strategies have been developed in recent years~\citep{ilharco2022editing,matena2022merging,jin2022dataless,yadav2023ties,bandarkar2024layer}.
Recently, several studies have applied model merging to enhance the capabilities of multimodal large language models. For example, REMEDY~\citep{zhu2025remedy} improves multitask performance and zero-shot generalization in VQA tasks. \citep{akiba2025evolutionary} enhance Japanese language understanding and generation, while \cite{chen2025bring} improve mathematical reasoning abilities. \cite{li2025transferringtextualpreferencesvisionlanguage} enable textual preference transfer by integrating a text-based reward model into an MLLM, without additional training.
Our study demonstrates that model merging can effectively endow MLLMs with strong abilities in multimodal code understanding and generation.

\section{Conclusion}
In conclusion, we have presented VisCodex, a unified multimodal framework that effectively integrates advanced visual comprehension with sophisticated code-generation capabilities through a novel task vector-based model merging strategy. By leveraging this efficient approach, VisCodex significantly enhances multimodal large language models without incurring the costs associated with full-scale retraining. We also introduced the Multimodal Coding Dataset (\ourdata{}), a comprehensive resource comprising 598k diverse, high-quality instruction-tuning examples, along with InfiBench-V, a rigorous benchmark designed specifically for realistic multimodal coding assessments. Extensive experiments confirm that VisCodex establishes a new state-of-the-art performance among open-source multimodal code generators, demonstrating capabilities competitive with leading proprietary models such as GPT-4o.

\hfill
\clearpage
\bibliographystyle{alpha}
\bibliography{main}

\hfill
\clearpage
\appendix


\section{Additional analysis experiments}
\label{app:exp_generalize}
\subsection{Generalizability of \ourdata{}}
To assess the generalizability of our dataset \ourdata{}, we conducted supervised fine-tuning experiments on two strong open-source MLLMs: \textbf{InternVL3-8B}~\citep{zhu2025internvl3} and \textbf{llava-llama3.1-8b}\footnote{\url{https://huggingface.co/modelscope/llava-llama3.1-8b}}~\citep{zhang2024mmllmsrecentadvancesmultimodal}. We evaluated the models on multiple established benchmarks, as shown in Table~\ref{tab:generalizability_internvl_final}. The results demonstrate that fine-tuning with \ourdata{} consistently and significantly enhances the multimodal coding abilities of both base models. In particular, we observe substantial improvements in both the \textbf{Design2Code} and \textbf{ChartMimic} tasks, as well as noticeable gains on the \textbf{MMCode} benchmark. These findings validate the robustness and strong transferability of \ourdata{} across different model architectures and suggest its value as a general-purpose resource for advancing multimodal code generation.

\begin{table}[h]
\centering
\resizebox{0.7\linewidth}{!}{%
\begin{tabular}{l|cc|cc|c}
\toprule
\textbf{Model} & \multicolumn{2}{c|}{\textbf{Design2Code}} & \multicolumn{2}{c|}{\textbf{ChartMimic}} & \textbf{MMCode} \\
& Low & High & Low & High & pass@1 \\
\midrule
InternVL3-8B & 85.3 & 87.6 & 43.1 & 46.6 & 6.8 \\
\textbf{InternVL3-8B-SFT} & \textbf{88.2} & \textbf{89.9} & \textbf{72.6} & \textbf{70.4} & \textbf{7.6} \\
\midrule
llava-llama3.1-8b & 7.3 & 78.4 & 6.2 & 4.8 & 2.3 \\
\textbf{llava-llama3.1-8b-SFT} & \textbf{82.8} & \textbf{90.5} & \textbf{70.7} & \textbf{68.2} & \textbf{4.2} \\
\bottomrule
\end{tabular}%
}
\caption{Generalization performance of \ourdata{}: Results of supervised fine-tuning on InternVL3-8B and llava-llama3.1-8b across multiple multimodal coding benchmarks.}
\label{tab:generalizability_internvl_final}
\end{table}

\subsection{Generality of Code Model Merging}
To verify that our model merging strategy is not limited to LLMs with Qwen2.5 backbones, we further conduct experiments using llava-llama3.1-8b~\citep{zhang2024mmllmsrecentadvancesmultimodal}, an MLLM based on the Llama-3.1~\citep{dubey2024llama} architecture. Specifically, we merge llava-llama3.1-8b with the code task vector ($\tau_{code}$) obtained from DeepSeek-R1-Distill-Llama-8B~\citep{deepseekai2025deepseekr1incentivizingreasoningcapability}, following the same procedure as in our previous Qwen2.5VL experiments. As shown in Table~\ref{tab:llava_merge}, the merged model consistently outperforms the non-merged baseline across all multimodal coding tasks. The performance improvement is particularly significant on the MMCode benchmark, where pass@1 increases from 4.2 to 6.5. These results, which are consistent with our findings on Qwen2.5VL, demonstrate the general effectiveness and broad applicability of the code model merging strategy.

\begin{table}[h]
\centering
\resizebox{0.7\linewidth}{!}{%
\begin{tabular}{l|cc|cc|c}
\toprule
\textbf{Method} & \multicolumn{2}{c|}{\textbf{Design2Code}} & \multicolumn{2}{c|}{\textbf{ChartMimic}} & \textbf{MMCode} \\
& Low-L & High-L & Low-L & High-L & pass@1 \\
\midrule
w/o model merge & 82.8 & 90.5 & 70.7 & 68.2 & 4.2 \\
w model merge & \textbf{83.3} & \textbf{90.7} & \textbf{70.9} & \textbf{68.8} & \textbf{6.5} \\
\bottomrule
\end{tabular}%
}
\caption{Ablation on model merging for \textbf{llava-llama3.1-8b} and \textbf{DeepSeek-R1-Distill-Llama-8B}. Results compare models with and without the merging strategy.}
\label{tab:llava_merge}
\end{table}

\section{Training Parameters and Training Cost}
\label{app:hyperparameters}
All models are trained on our instruction-tuning dataset of 598K examples from \ourdata{}. We employ the AdamW optimizer with a 10\% linear warm-up followed by a cosine learning rate decay. The maximum learning rate is set to $1 \times 10^{-5}$, with a batch size of 128 and a maximum sequence length of 8K tokens. Training the 8B model for two epochs takes approximately 16 hours on 8 nodes, each equipped with 8$\times$A100 GPUs with 40~GB VRAMs. Training the 33B model under the same setup takes approximately 2 days.


\section{Human Evaluation}

To assess the alignment between automatic evaluation and human expert judgment, we randomly sampled 100 questions from InfiBench-V and selected five MLLMs for evaluation: GPT-4o-mini, \ourmodel{}-33B, \ourmodel{}-8B, Qwen2.5-VL-7B-Instruct, and Qwen2.5-VL-32B-Instruct. Each model was tasked with generating responses to all sampled questions, resulting in a total of 500 model outputs.

\paragraph{Annotator Details and Protocol.}
We employed three annotators, all of whom are graduate students in computer science with strong domain expertise relevant to the benchmark tasks. Although all annotators are non-native English speakers, they are proficient in English and familiar with the style and requirements of the benchmark questions.

Before the main evaluation, the annotators underwent a comprehensive training and calibration process, including practice with a subset of the dataset, review of the evaluation criteria, and targeted feedback to ensure consistent understanding of the annotation standards. A final readiness test was conducted to confirm alignment among annotators prior to the main evaluation phase.

During evaluation, all model-generated responses were anonymized and presented in a standardized format, removing all identifying information about the originating model. Annotators independently evaluated the samples without communication or influence from others, ensuring impartiality throughout the process.

For each question, the annotators were provided with the question, the model-generated answer, and the accepted StackOverflow answer as a reference. Each annotator was asked to indicate whether the model's answer correctly solved the problem (1 for correct, 0 for incorrect).

\paragraph{Automatic Evaluation.}
For the automatic InfiBench-V evaluation, we adopted the same metrics as described in the main text, assigning a score in the range 0--100 to each response. For the purpose of agreement analysis with human annotations, we further mapped these scores to binary ``pass'' or ``fail'' labels according to a defined threshold. This binarization enabled direct comparison between the automatic evaluation and the binary judgments provided by human annotators.

\paragraph{Agreement Analysis.}
Table~\ref{tab:human_label} reports the pairwise agreement ratios (as in MT-Bench~\citep{zheng2023judgingllmasajudgemtbenchchatbot}) among InfiBench-V and the human annotators, as well as between human annotators themselves. On average, InfiBench-V achieved an agreement rate of 86.07\% with human annotators, which is higher than the inter-annotator agreement (79.33\%). Notably, the agreement rate of InfiBench-V is comparable to that reported in previous works~\citep{li2024infibench} (85.1\%), demonstrating the reliability of our evaluation method.


\begin{table}[ht]
    \centering
    \resizebox{0.6\linewidth}{!}{
        \begin{tabular}{lcccc}
            \toprule
            Judge & A-1 & A-2 & A-3 & Average \\
            \midrule
            InfiBench-V & 87.00\% & 87.00\% & 84.20\% & 86.07\% \\
            A-1         &   --    & 80.80\% & 79.20\% & 80.00\% \\
            A-2         & 80.80\% &   --    & 78.00\% & 79.40\% \\
            A-3         & 79.20\% & 78.00\% &   --    & 78.60\% \\
            \bottomrule
        \end{tabular}
    }
    \caption{Agreement ratios between InfiBench-V and human annotators, as well as between annotators.}
    \label{tab:human_label}
\end{table}

\section{Data Statistics}

\subsection{Data Statistics of \ourdata{}}
\label{app:mcd_data_statistics}
To ensure a fair evaluation and prevent data contamination, we performed deduplication between our dataset and the evaluation benchmarks using SimHash~\citep{manku2007detecting}, removing any data with a similarity score greater than 0.9 to the benchmark test samples. After deduplication, the resulting dataset contains a total of 598k examples across four domains: HTML, Chart, QA, and Algorithm. Table~\ref{tab:mcd_distribution} presents the statistics of the dataset, including the number of examples and the average token length (with standard deviation) for each domain.

\begin{table}[h]
    \centering
    
    \resizebox{0.7\linewidth}{!}{
    \renewcommand\tabcolsep{6.0pt}
    \begin{tabular}{lcccc}
    \toprule
        \textbf{Data} & \textbf{HTML}  & \textbf{Chart} & \textbf{QA} & \textbf{Algorithm}  \\ 
        \midrule
        Size & 200k & 210k & 59k & 129k \\
        Avg Length & 632±144  & 551±190 & 1022±776  & 969±321 \\
    \bottomrule
    \end{tabular}
    }
    \caption{Statistics of the \ourdata{} by domain. For each domain, the number of samples and the average token length (mean ± standard deviation) are reported.}
    \label{tab:mcd_distribution}
\end{table}

\subsection{Distribution of selected benchmark questions in InfiBench-V}
\label{app:infibench_v_statistics}
Table~\ref{tab:infibenchv-distribution} presents the detailed distribution of questions in InfiBench-V across various programming domains and languages. The benchmark consists of five main categories: Front-End, Back-End, Data Science \& Machine Learning (DS \& ML), IT Operations (IT Ops), and Mobile \& Desktop development, covering a total of 13 programming languages and 322 carefully curated questions.

\begin{table}[h]
    \centering
    \resizebox{0.7\linewidth}{!}{
        \begin{tabular}{llrr}
            \toprule
            \textbf{Category} & \textbf{Language} & \textbf{Count} & \textbf{Category Total} \\
            \midrule
            \multirow{3}{*}{Front-End} & CSS & 30 & \multirow{3}{*}{100} \\
            & HTML & 39 & \\
            & JavaScript & 31 & \\
            \midrule
            \multirow{6}{*}{Back-End} & Java & 30 & \multirow{6}{*}{75} \\
            & PHP & 24 & \\
            & Go & 5 & \\
            & Ruby & 5 & \\
            & Rust & 6 & \\
            & C++ & 5 & \\
            \midrule
            \multirow{2}{*}{DS \& ML} & Python & 90 & \multirow{2}{*}{95} \\
            & R & 5 & \\
            \midrule
            IT Ops & Bash & 11 & 11 \\
            \midrule
            Mobile \& Desktop & Dart & 41 & 41 \\
            \midrule
            \textbf{Total} & & \textbf{322} & \textbf{322} \\
            \bottomrule
        \end{tabular}
    }
    \caption{Distribution of InfiBench-V questions across programming domains and languages.}
    \label{tab:infibenchv-distribution}
\end{table}

\clearpage
\section{Prompt}

\begin{tcolorbox}[
    colback=white,
    colframe=gray,
    title=\textbf{Prompt Template for GPT-4o Automatic Evaluation on InfiBench-V},
    fonttitle=\bfseries\large,
    arc=4mm,
    breakable
]

You are a professional code assistant evaluation expert. Your task is to assess the quality of a model-generated answer to a programming-related question by comparing it with the gold reference answer. The question includes both text and a screenshot image (which may contain code, error messages, or UI context). You must consider both modalities when evaluating the answer.
\\

Please first read the \textbf{question (text + image)}, then read the \textbf{model-generated answer}, and compare it carefully with the \textbf{reference (gold) answer}.\\

You must provide clear and detailed justifications before assigning scores. The scoring dimensions are weighted as follows: Correctness (50 points) and Completeness (50 points).\\

Below is a programming question-answering example.\\

\textbf{[Question]}  
\{question\}\\

\textbf{[Note]}  
The screenshot image provided alongside this question is part of the question context. You must use it to fully understand the problem being asked.

\textbf{[Screenshot Image]}  \\
\{base64\_image\}\\

\textbf{[Model Answer]} \\ 
\{model\_answer\} \\

\textbf{[Reference Answer]}  \\
\{reference\_answer\} \\

Please evaluate the model's answer based on the following two dimensions: \\

1. \textbf{Correctness (0--50)}: Is the answer factually accurate and technically correct? Does it address the problem without errors or misleading information?  \\
2. \textbf{Completeness (0--50)}: Does the answer cover all critical elements present in the reference answer? Are any key details missing? \\

\medskip
\textbf{---}

\textbf{Justification:}  \\
1. Correctness: \textless Your reasoning here\textgreater  \\
2. Completeness: \textless Your reasoning here\textgreater \\

\medskip

\textbf{Scores:}
\begin{itemize}
    \item Correctness: X/50
    \item Completeness: X/50
\end{itemize}

\textbf{Total Score: X/100}

\end{tcolorbox}

\newpage
\section{Additional Examples}
\label{app:additional_cases}
\subsection{MMCode Case 1}

\lstset{
  language=Python,
  basicstyle=\ttfamily\scriptsize, 
  keywordstyle=\color{blue},
  stringstyle=\color{orange},
  commentstyle=\color{gray},
  showstringspaces=false,
  numberstyle=\tiny,
  breaklines=true,        
  breakatwhitespace=true, 
  frame=none,
  columns=flexible         
}

\begin{figure}[H]
\centering
\scriptsize
\begin{tcolorbox}[
  enhanced,
  sharp corners,
  colback=white,
  colframe=black,
  left=1pt, right=1pt, top=1pt, bottom=1pt,
  boxsep=2pt,
]
\scriptsize
\begin{minipage}{0.68\linewidth}
\textbf{Question:} Alice has a binary string. She thinks a binary string is beautiful if and only if it doesn't contain the substring $\text{"010"}$. 

In one step, Alice can change a $0$ to a $1$ or vice versa. Count and print the minimum number of steps needed to make Alice see the string as beautiful.

\textbf{Example}

$b=\textbf{010}$

She can change any one element and have a beautiful string.














\textbf{Sample Input 0}

\begin{tabular}{ll}
\texttt{STDIN} & \texttt{Function} \\
\texttt{-----} & \texttt{--------} \\
7 & length of string n = 7 \\
0101010 & b = '0101010' \\
\end{tabular}

\textbf{Sample Output 0}

2

\textbf{Explanation 0:}

In this sample, $b=\text{"0101010"}$

The figure on the right shows a way to get rid of each instance of $\text{"010"}$:
Make the string beautiful by changing $2$ characters ($b[2]$ and $b[5]$).
\end{minipage}%
\hfill
\begin{minipage}{0.28\linewidth}
\centering
\includegraphics[width=0.95\linewidth]{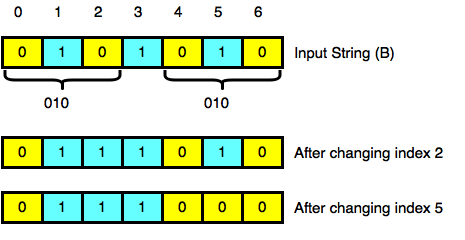} 





\end{minipage}
\end{tcolorbox}

\begin{tcolorbox}[
  enhanced,
  sharp corners,
  colback=white,
  colframe=black,
  left=1pt, right=1pt, top=1pt, bottom=1pt,
  boxsep=0pt,
]


\textcolor{blue}{\textbf{Qwen2.5-VL-7B-Instruct}}: 

\begin{lstlisting}
def beautifulBinaryString(b):
	b = list(b)
	count = 0
	for i in range(len(b) - 2):
		if b[i] == '0' and b[i + 1] == '1' and b[i + 2] == '0':
			b[i] = '1'
			b[i + 1] = '0'
			count += 2
	return count
\end{lstlisting} \xmark

\centering
\begin{tcolorbox}[
  enhanced,
  sharp corners,
  colback=gray!10,
  colframe=gray!10,
  boxrule=0pt,
  boxsep=1pt,
  left=0pt, right=0pt, top=0pt, bottom=0pt,
  width=\dimexpr\linewidth+1pt\relax 
]

\textcolor{red}{\textbf{\ourmodel{}-8B}}: 

To solve the problem of making a binary string ``beautiful'' (i.e., removing all occurrences of the substring ``010''), we need to identify and count these substrings within the given binary string. For each occurrence of ``010'', we will determine the minimal changes required to eliminate it.

Here's how we can approach the problem:

1. \textbf{Identify Occurrences}: Traverse through the string and find all instances of the substring ``010''. \\
2. \textbf{Count Changes}: For each found ``010'', determine the minimal number of changes required to break this pattern. Typically, changing either the middle `1` to `0` or the last `0` to `1` suffices. \\
3. \textbf{Sum Up Changes}: Sum up the changes for all identified ``010'' substrings to get the total number of steps required. \\

Let's implement this logic in Python:

\begin{lstlisting}
def beautifulBinaryString(b):
    # Initialize the counter for the number of changes
    changes = 0
    
    # Iterate over the string to find all occurrences of "010"
    i = 0
    while i <= len(b) - 3:
        if b[i:i+3] == "010":
            # We found a "010" pattern, increment the changes counter
            changes += 1
            # Move the index past this pattern
            i += 3
        else:
            # Move to the next character
            i += 1
            
    return changes

\end{lstlisting}

    
            
\cmark

\end{tcolorbox}
\end{tcolorbox}

\caption{Examples of MMCode problems that are successfully solved by VisCodex-8B but not by Qwen2.5-VL-7B-Instruct. }
\label{fig:case_mmcode1}
\vspace{-4mm}
\end{figure}

\subsection{MMCode Case 2}
\begin{figure}[h]
\centering
\scriptsize
\begin{tcolorbox}[
  enhanced,
  sharp corners,
  colback=white,
  colframe=black,
  left=1pt, right=1pt, top=1pt, bottom=1pt,
  boxsep=2pt,
]
\scriptsize
\begin{minipage}{0.68\linewidth}
\textbf{Question:}  
You have most definitely heard the legend of King Arthur and the Knights of the Round Table. Almost all versions of this story proudly point out that the roundness of the Round Table is closely related to Arthur’s belief of equality among the Knights. That is a lie! In fact, Arthur’s choice of table is conditioned by his childhood traumas.

In fact, Arthur was forced to clean up quadratic tables from a young age after a tournament in pick-up sticks$^{1}$ had been played on them. After the tournament, typically there would be a bunch of sticks on the table that do not touch each other. In the spirit of the game, the organizers issued strict regulations for the table cleaners. More precisely, the sticks on the table need to be removed one by one in a way that the cleaners pull them in the shortest way towards the edge of the table closest to where they are currently sitting. They also mustn’t rotate or touch the other sticks while doing this (not even in the edge points).

In this task, we will represent the table in the coordinate system with a square that has opposite points in the coordinates $(0,0)$ and $(10\,000, 10\,000)$, whereas the sticks will be represented with straight line segments that lie within that square. We will assume that Arthur is sitting at the edge of the table lying on the $x$-axis. Then the movement of the stick comes down to translating the line segment along the shortest path towards the $x$-axis until the stick falls off the table (as shown in the right image). It is your task to help Arthur determine the order of stick movements that meets the requirements from the previous paragraph.
\end{minipage}%
\hfill
\begin{minipage}{0.28\linewidth}
\centering
\includegraphics[width=0.95\linewidth]{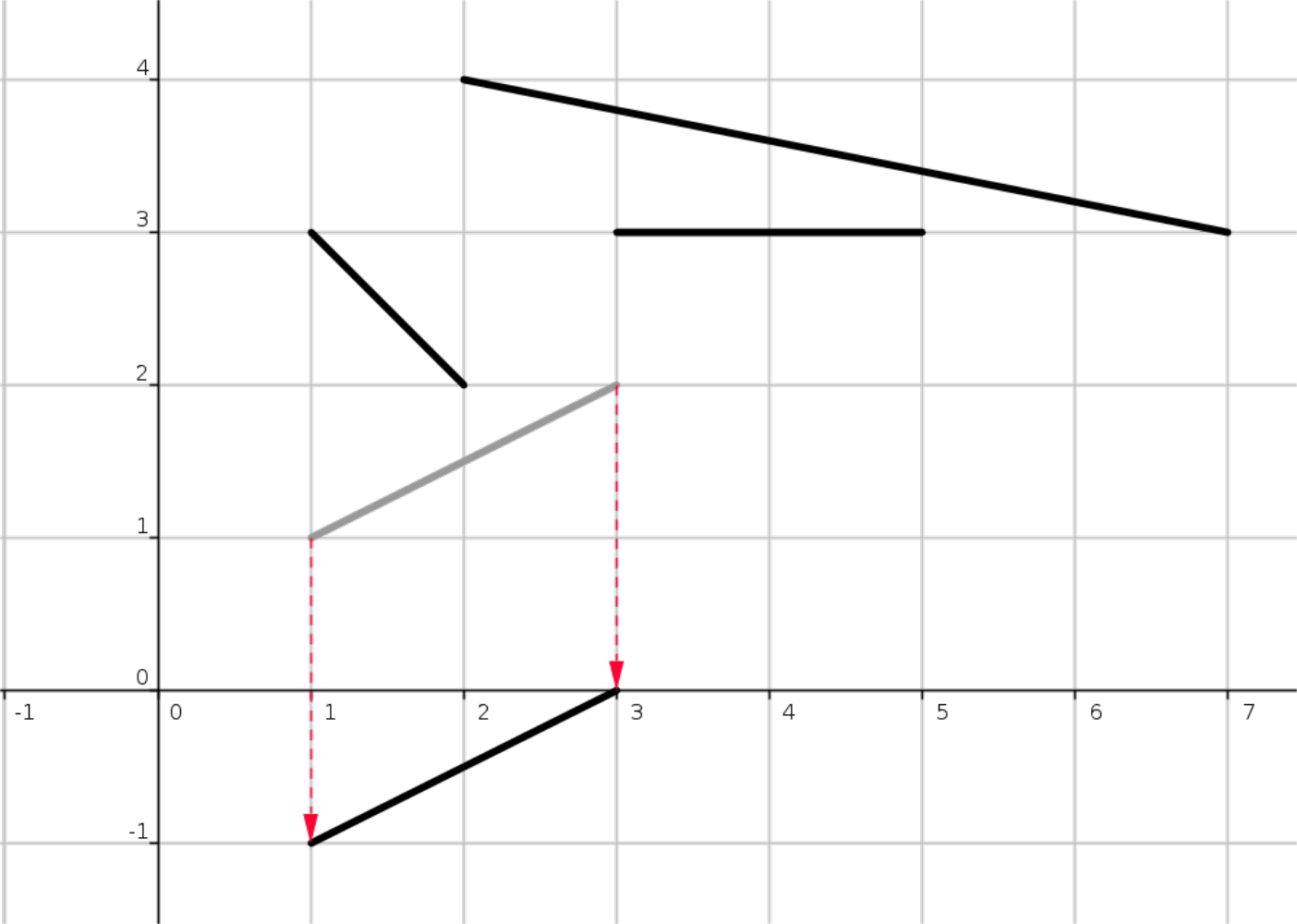} 
\end{minipage}
\end{tcolorbox}

\begin{tcolorbox}[
  enhanced,
  sharp corners,
  colback=white,
  colframe=black,
  left=1pt, right=1pt, top=1pt, bottom=1pt,
  boxsep=0pt,
]


\textcolor{blue}{\textbf{Qwen2.5-VL-7B-Instruct}}: 

\begin{lstlisting}
import sys
input = sys.stdin.readline

def solve():
	n = int(input())
	ls = []
	for _ in range(n):
		(x1, y1, x2, y2) = map(int, input().split())
		ls.append((x1, y1, x2, y2))
	ls.sort(key=lambda x: (x[1] - x[3], x[0]))
	ans = []
	for (x1, y1, x2, y2) in ls:
		if y1 == y2:
			ans.append(1)
		else:
			ans.append(2)
	print(*ans)
solve()
\end{lstlisting} \xmark

\centering
\begin{tcolorbox}[
  enhanced,
  sharp corners,
  colback=gray!10,
  colframe=gray!10,
  boxrule=0pt,
  boxsep=1pt,
  left=0pt, right=0pt, top=0pt, bottom=0pt,
  width=\dimexpr\linewidth+1pt\relax 
]

\textcolor{red}{\textbf{\ourmodel{}-8B}}: 

\begin{lstlisting}
def remove_sticks_order(n, sticks):
    # Create a list of tuples where each tuple is (index, x1, y1, x2, y2)
    indexed_sticks = [(i + 1, x1, y1, x2, y2) for i, (x1, y1, x2, y2) in enumerate(sticks)]
    
    # Sort the sticks based on the y-coordinate in descending order
    indexed_sticks.sort(key=lambda x: (-x[2], -x[4]))
    
    # Extract the indices of the sorted sticks
    sorted_indices = [stick[0] for stick in indexed_sticks]
    
    return sorted_indices

# Example usage:
if __name__ == "__main__":
    import sys
    input = sys.stdin.read
    data = input().split()
    
    n = int(data[0])
    sticks = []
    index = 1
    for _ in range(n):
        x1 = int(data[index])
        y1 = int(data[index + 1])
        x2 = int(data[index + 2])
        y2 = int(data[index + 3])
        sticks.append((x1, y1, x2, y2))
        index += 4
    
    result = remove_sticks_order(n, sticks)
    print(" ".join(map(str, result)))
\end{lstlisting}

\cmark

\end{tcolorbox}
\end{tcolorbox}

\caption{Examples of MMCode problems that are successfully solved by VisCodex-8B but not by Qwen2.5-VL-7B-Instruct. }
\label{fig:casemmcode2}
\end{figure}

\clearpage

\subsection{InfiBench-V Case 1}
\begin{figure}[h]
\centering
\scriptsize
\begin{tcolorbox}[
  enhanced,
  sharp corners,
  colback=white,
  colframe=black,
  left=1pt, right=1pt, top=1pt, bottom=1pt,
  boxsep=2pt,
  width=\linewidth
]
\scriptsize

\begin{minipage}{0.68\linewidth}
\textbf{Question:}  
I attempted to change the colors in my macOS Terminal by following a guide and modifying my \texttt{.bash\_profile} file. After creating and saving the \texttt{.bash\_profile}, every time I open the Terminal, I encounter the following errors as shown in the screenshot on the right:

Despite deleting the \texttt{.bash\_profile}, the issue persists. Every command I enter is followed by these errors:

\begin{lstlisting}[language=bash]
-bash:    git: command not found
-bash: gt: command not found
-bash: /dev/null: Permission denied
\end{lstlisting}

Here is the content I added to the \texttt{.bash\_profile} file:

\begin{lstlisting}[language=bash]
export CLICOLOR=1
export LSCOLORS=GxBxCxDxexegedabagaced
parse_git_branch() {
    git branch 2> /dev/null |
    sed -e '/^[^*]/d' -e 's/* (.*)/ (\1)/'
}
export PS1="\e[0;35m->> \e[1;34m\W\e[0;32m\$(parse_git_branch)\e[0;37m $ "
\end{lstlisting}

Please help me resolve these issues and understand why these errors are occurring.
\end{minipage}%
\hfill
\begin{minipage}{0.28\linewidth}
    \centering
    \vspace{0pt}
    \includegraphics[width=0.95\linewidth]{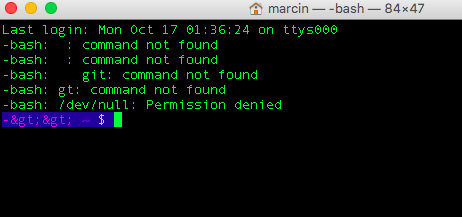} 
\end{minipage}
\end{tcolorbox}










\begin{tcolorbox}[
  enhanced,
  sharp corners,
  colback=white,
  colframe=black,
  left=1pt, right=1pt, top=1pt, bottom=1pt,
  boxsep=0pt,
]
\scriptsize


\textcolor{blue}{\textbf{Response}}: 

{\tiny
The errors you're encountering are likely due to incorrect syntax in your \texttt{.bash\_profile} file. Let's address each issue:

\begin{enumerate}[left=1em, itemsep=-0.2em]
\item \textbf{Incorrect Redirection Symbol}: In your \texttt{parse\_git\_branch()} function, you used \verb|&gt;| instead of \verb|>| for redirecting output to \texttt{/dev/null}. The correct line should be:
\begin{lstlisting}[language=bash]
git branch 2> /dev/null |
\end{lstlisting}

\item \textbf{Extra Spaces in PS1}: Your \texttt{PS1} variable has some extra spaces that might cause issues. It should look like this:
\begin{lstlisting}[language=bash]
export PS1="\e[0;35m&gt;&gt; \e[1;34m\W\e[0;32m\$(parse_git_branch)\e[0;37m $ "
\end{lstlisting}
Make sure there are no unintended spaces around special characters or within the escape sequences.

\item \textbf{Check for Hidden Characters}: Sometimes, copy-pasting from certain editors or websites can introduce hidden characters. Ensure your \texttt{.bash\_profile} doesn't contain any non-printable characters. You can use a text editor that shows whitespace or use a command-line tool to check for them.

\item \textbf{File Permissions}: Ensure your \texttt{.bash\_profile} file has the correct permissions. You can set it with:
\begin{lstlisting}[language=bash]
chmod 644 ~/.bash_profile
\end{lstlisting}

\item \textbf{Recreate .bash\_profile}: If the problem persists, try recreating the \texttt{.bash\_profile} file manually without copying and pasting. Open a new terminal and type the commands directly into a new file.

\end{enumerate}

After making these changes, save the file and restart your terminal session. This should resolve the errors you're seeing. If the problem continues, double-check for any other syntax errors or unexpected characters in the file.
}

\centering
\begin{tcolorbox}[
  enhanced,
  sharp corners,
  colback=gray!10,
  colframe=gray!10,
  boxrule=0pt,
  boxsep=1pt,
  left=0pt, right=0pt, top=0pt, bottom=0pt,
  width=\dimexpr\linewidth+1pt\relax 
]


\textbf{Category: IT Ops——Bash}

\textbf{Judge: GPT-4o}

{\tiny
\begin{enumerate}[left=1em, itemsep=0.05em]
    \item \textbf{Correctness:}
        \begin{itemize}[left=1em, itemsep=0.05em]
            \item The model answer correctly identifies the primary issue with the \texttt{.bash\_profile} file, which is the use of HTML entities (\verb|&gt;|) instead of the correct characters (\verb|>|). This aligns with the reference answer.
            \item The model also addresses the potential for non-printable characters and file permissions, which, while not mentioned explicitly in the reference answer, are reasonable considerations given the symptoms described.
            \item Additionally, the model suggests recreating the \texttt{.bash\_profile} file to ensure there are no hidden characters, which is a practical troubleshooting step that could help resolve the persistent issues.
            \item Overall, the model's suggestions are factually accurate and technically correct, directly addressing the syntax errors.
        \end{itemize}
    \item \textbf{Completeness:}
        \begin{itemize}[left=1em, itemsep=0.05em]
            \item The reference answer is concise and focuses solely on the issue of HTML entities, while the model answer provides a more comprehensive approach by including additional troubleshooting steps (checking for hidden characters, file permissions, and recreating the file).
            \item Although the model answer includes extra steps beyond the reference answer, it does not miss any critical elements present in the reference. Instead, it enhances the response with additional useful advice for resolving the issue thoroughly.
            \item The model's completeness is therefore high, as it covers the key issue identified in the reference answer and adds further depth.
        \end{itemize}
\end{enumerate}
\textbf{Scores:}
\begin{itemize}[left=1.5em, itemsep=0.2em]
    \item Correctness: 50/50
    \item Completeness: 50/50
\end{itemize}
}

\textbf{Total Score: 100/100}

\end{tcolorbox}
\end{tcolorbox}

\caption{A case of InfiBench-V from the IT Ops category.}
\label{fig:caseinfibench1}
\vspace{-4mm}
\end{figure}

\clearpage
\subsection{InfiBench-V Case 2}

\lstdefinelanguage{JavaScript}{
  keywords={break, case, catch, const, continue, debugger, default, delete, do, else, export, for, function, if, import, in, instanceof, let, new, return, super, switch, this, throw, try, typeof, var, void, while, with, yield, class, enum, await, implements, interface, package, private, protected, public, static, null, true, false},
  keywordstyle=\color{blue}\bfseries,
  ndkeywords={typeof, new, true, false, catch, function, return, null, catch, switch, var, if, in, while, do, else, case, break},
  ndkeywordstyle=\color{red}\bfseries,
  identifierstyle=\color{black},
  sensitive=false,
  comment=[l]{//},
  morecomment=[s]{/*}{*/},
  commentstyle=\color{gray}\ttfamily,
  stringstyle=\color{orange}\ttfamily,
  morestring=[b]',
  morestring=[b]",
}

\begin{figure}[h]
\centering
\scriptsize
\begin{tcolorbox}[
  enhanced,
  sharp corners,
  colback=white,
  colframe=black,
  left=1pt, right=1pt, top=1pt, bottom=1pt,
  boxsep=2pt,
  width=\linewidth
]
\scriptsize

\begin{minipage}{0.68\linewidth}
\textbf{Question:}  
I have a dataset with absolute numbers for various variants over time, and I want to create a plot similar to the one shown in the image, depicting the abundance of these variants over time. I am working in Python, and I would prefer to use matplotlib, although other libraries are also acceptable if necessary.

Does Python offer any specific function or tool for generating this type of plot? Additionally, what is the common name for this kind of plot, as I have encountered it before but am having trouble finding relevant resources online?
\end{minipage}%
\hfill
\begin{minipage}{0.28\linewidth}
    \centering
    \vspace{0pt}
    \includegraphics[width=0.95\linewidth]{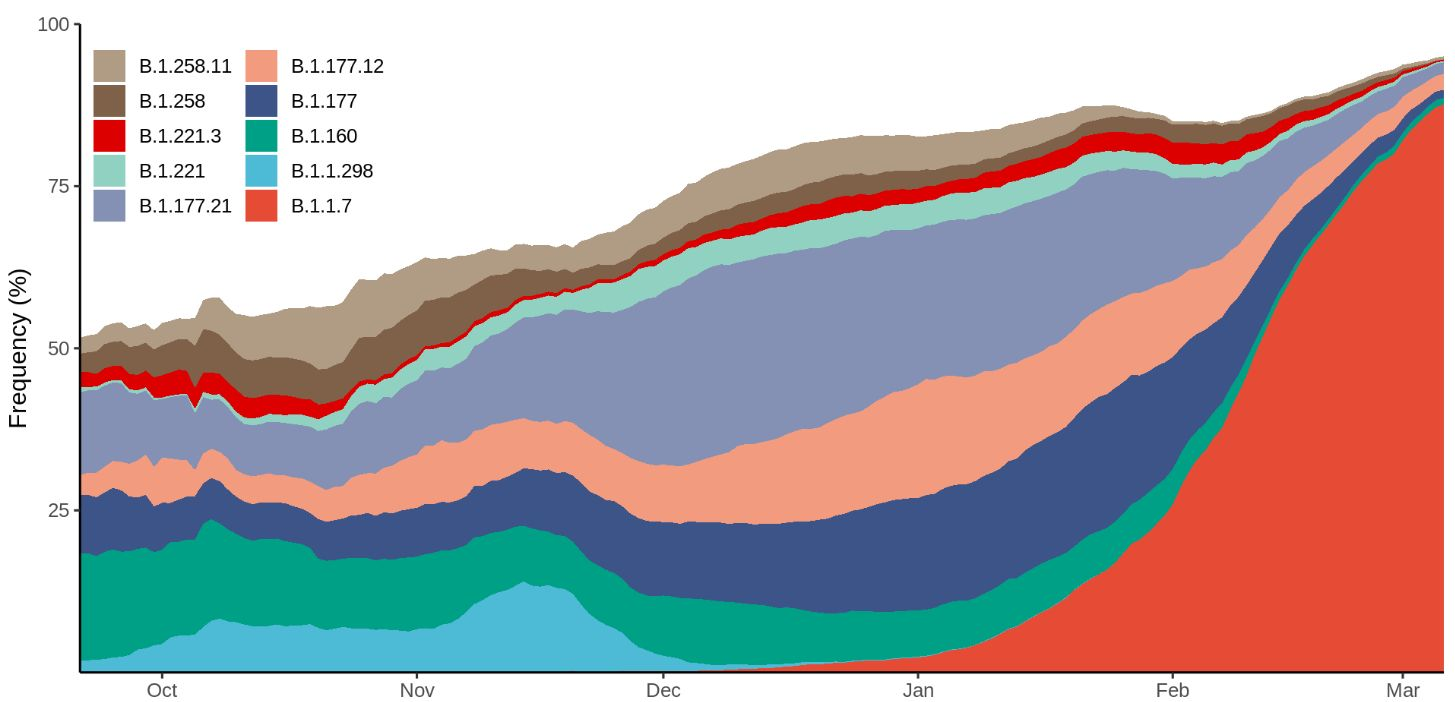} 
\end{minipage}

\end{tcolorbox}

\begin{tcolorbox}[
  enhanced,
  sharp corners,
  colback=white,
  colframe=black,
  left=1pt, right=1pt, top=1pt, bottom=1pt,
  boxsep=0pt,
]
\scriptsize


\textcolor{blue}{\textbf{Response}}: 
{\tiny
The type of plot shown in the image is commonly known as an "area plot" or "stacked area plot." In Python, you can create such plots using the \texttt{matplotlib} library. Although there isn't a specific function called "area plot" in \texttt{matplotlib}, you can use the \texttt{stackplot} function to generate this type of visualization.

Here's a basic example of how you can create a stacked area plot using \texttt{matplotlib}:

\begin{lstlisting}[language=Python, basicstyle=\tiny\ttfamily]
import matplotlib.pyplot as plt
import numpy as np

# Example data
dates = np.arange('2021-01', '2021-03', dtype='datetime64[D]')
variant1 = np.random.rand(len(dates))
variant2 = np.random.rand(len(dates))
variant3 = np.random.rand(len(dates))

# Normalize data to sum to 1 for each day
total = variant1 + variant2 + variant3
variant1 /= total
variant2 /= total
variant3 /= total

# Create stacked area plot
plt.stackplot(dates, variant1, variant2, variant3, labels=['Variant 1', 'Variant 2', 'Variant 3'], alpha=0.8)
plt.legend(loc='upper left')
plt.xlabel('Date')
plt.ylabel('Frequency (%)')
plt.title('Abundance of Variants Over Time')
plt.show()
\end{lstlisting}

In this example, \texttt{variant1}, \texttt{variant2}, and \texttt{variant3} are arrays representing the frequencies of different variants over time. You should replace these with your actual data. Make sure your data is normalized (i.e., the sum of all variants for each time point should equal 1, or 100\% if expressed in percentage terms) before plotting.

If you prefer using \texttt{pandas}, you can also leverage its capabilities to simplify data manipulation and plotting:

\begin{lstlisting}[language=Python, basicstyle=\tiny\ttfamily]
import pandas as pd
import matplotlib.pyplot as plt

# Example data
data = {
    'Date': pd.date_range(start='2021-01-01', periods=60, freq='D'),
    'Variant1': np.random.rand(60),
    'Variant2': np.random.rand(60),
    'Variant3': np.random.rand(60)
}

df = pd.DataFrame(data)
df.set_index('Date', inplace=True)

# Normalize data to sum to 1 for each day
df = df.div(df.sum(axis=1), axis=0)

# Plot
df.plot.area(alpha=0.8)
plt.ylabel('Frequency (%)')
plt.title('Abundance of Variants Over Time')
plt.show()
\end{lstlisting}

This approach simplifies the handling of data, especially if it's already in a \texttt{pandas} DataFrame format. Adjust the data and labels as needed for your specific dataset.
}

\centering
\begin{tcolorbox}[
  enhanced,
  sharp corners,
  colback=gray!10,
  colframe=gray!10,
  boxrule=0pt,
  boxsep=1pt,
  left=0pt, right=0pt, top=0pt, bottom=0pt,
  width=\dimexpr\linewidth+1pt\relax 
]


\textbf{Category: DS \& ML——Python}

\textbf{Evaluation Criteria: Keyword Matching \& GPT-4o}

{\tiny
\textbf{Keywords:} \texttt{stackplot}, \texttt{matplotlib} \\
\textbf{Keyword Matching Score:} \texttt{100} \\
\textbf{GPT-4o Score:} \texttt{100} \\
}

\textbf{Total Score: 100}

\end{tcolorbox}
\end{tcolorbox}

\caption{A case of InfiBench-V from the DS \& ML category. }
\label{fig:infibenc2}
\vspace{-4mm}
\end{figure}

\clearpage
\subsection{InfiBench-V Case 3}

\begin{figure}[h]
\centering
\scriptsize
\begin{tcolorbox}[
  enhanced,
  sharp corners,
  colback=white,
  colframe=black,
  left=1pt, right=1pt, top=1pt, bottom=1pt,
  boxsep=2pt,
  width=\linewidth
]
\scriptsize

\begin{minipage}{0.68\linewidth}
\textbf{Question:}  
I am working on a 2D game in JavaScript and I am encountering a trigonometry issue. The game involves a player shooting at a target, but there is an obstacle blocking the shot between the player and the target. I need to find the coordinates of this obstacle.

Here is the information I have:
\begin{enumerate}
    \item The starting coordinates of the shot are \((x_1, y_1)\).
    \item The ending coordinates of the shot are \((x_2, y_2)\).
    \item The distance from the starting coordinates to the obstacle is known.
\end{enumerate}

How can I calculate the \((x, y)\) coordinates of the obstacle along the line of the shot? Additionally, I would appreciate an efficient solution implemented in JavaScript.

Please refer to the attached image for a visual representation of the problem.
\end{minipage}%
\hfill
\begin{minipage}{0.28\linewidth}
    \centering
    \vspace{0pt}
    \includegraphics[width=0.95\linewidth]{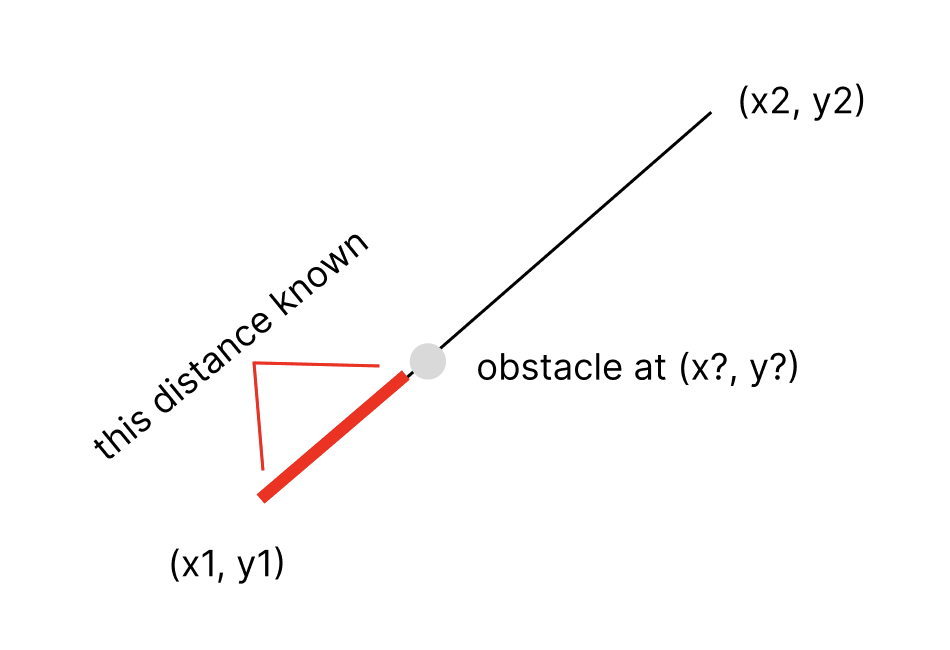} 
\end{minipage}

\end{tcolorbox}

\begin{tcolorbox}[
  enhanced,
  sharp corners,
  colback=white,
  colframe=black,
  left=1pt, right=1pt, top=1pt, bottom=1pt,
  boxsep=0pt,
]
\scriptsize


\textcolor{blue}{\textbf{Response}}: 

{\tiny
To solve the problem of finding the coordinates of the obstacle along the line of the shot, you can use vector mathematics. Given the starting point \((x1, y1)\), the ending point \((x2, y2)\), and the distance \(d\) from the starting point to the obstacle, you can calculate the coordinates of the obstacle as follows:

\begin{enumerate}[left=1em, itemsep=-0.4em, topsep=0pt]
    \item Calculate the direction vector from \((x_1, y_1)\) to \((x_2, y_2)\):
    \[
    \text{direction} = (x_2 - x_1,\, y_2 - y_1)
    \]
    \item Normalize the direction vector to get the unit vector in the direction of the shot:
    \[
    \text{magnitude} = \sqrt{(x_2 - x_1)^2 + (y_2 - y_1)^2}
    \]
    \[
    \text{unit\_direction} = \left(\frac{x_2 - x_1}{\text{magnitude}},\, \frac{y_2 - y_1}{\text{magnitude}}\right)
    \]
    \item Multiply the unit direction vector by the distance \(d\) to find the displacement vector from the starting point to the obstacle:
    \[
    \text{displacement} = (d \times \text{unit\_direction}_x,\, d \times \text{unit\_direction}_y)
    \]
    \item Add the displacement vector to the starting point to get the coordinates of the obstacle:
    \[
    \text{obstacle}_x = x_1 + \text{displacement}_x
    \]
    \[
    \text{obstacle}_y = y_1 + \text{displacement}_y
    \]
\end{enumerate}

Here's how you can implement this in JavaScript:

\begin{lstlisting}[language=JavaScript, basicstyle=\tiny\ttfamily]

\end{lstlisting}

This code will give you the coordinates of the obstacle along the line of the shot, given the specified distance from the starting point.
}

\centering
\begin{tcolorbox}[
  enhanced,
  sharp corners,
  colback=gray!10,
  colframe=gray!10,
  boxrule=0pt,
  boxsep=1pt,
  left=0pt, right=0pt, top=0pt, bottom=0pt,
  width=\dimexpr\linewidth+1pt\relax 
]

\textbf{Category: Front-End——JavaScript}

\textbf{Evaluation Criteria: Unit Testing}

{\tiny
\textbf{Sample Test Case:} \\
\texttt{input:    \{"x1":0.0,"y1":0.0,"x2":10.0,"y2":0.0,"distance":5.0\},} \\
\texttt{expected: [5.0, 0.0],} \\
\texttt{actual:   [5.0, 0.0],\ ...} \\
\textcolor{gray}{(only partial cases shown)}
}

\textbf{Total Score: 100}

\end{tcolorbox}
\end{tcolorbox}

\caption{A case of InfiBench-V from the Front-End category. }
\label{fig:caseinfibench3}
\vspace{-4mm}
\end{figure}

\lstdefinelanguage{Dart}{
  morekeywords={abstract,else,import,show,as,enum,in,static,assert,export,interface,super,async,extends,is,switch,await,extension,late,sync,break,external,library,this,case,factory,mixin,throw,catch,false,new,true,class,final,null,try,const,finally,on,typedef,continue,for,operator,var,covariant,Function,part,void,default,get,rethrow,while,deferred,hide,return,with,do,if,set,yield,dynamic},
  sensitive=true,
  morecomment=[l]{//},
  morecomment=[s]{/*}{*/},
  morestring=[b]",
  morestring=[b]'
}

\clearpage

\subsection{InfiBench-V Case 4}

\begin{figure}[H]
\centering
\scriptsize
\begin{tcolorbox}[
  enhanced,
  sharp corners,
  colback=white,
  colframe=black,
  left=1pt, right=1pt, top=1pt, bottom=1pt,
  boxsep=2pt,
  width=\linewidth
]
\scriptsize

\begin{minipage}{0.68\linewidth}
\textbf{Question:}  
{ \tiny
I am developing a Flutter application where I want to draw a square on the screen at the location of a click. The square should be centered at the click point, but currently, it is being drawn at seemingly random positions. I am using `event.pointerEvent.localPosition.dx` and `event.pointerEvent.localPosition.dy` to obtain the click coordinates, and I am passing these coordinates to `Rect.fromCenter` to draw the square.

Below is my Flutter code, which includes the `SpriteSquare` widget and the `handleAdd` function from its parent widget, `MySpriteGame`.

\textbf{sprite\_square.dart}
\begin{lstlisting}[language=Dart, basicstyle=\tiny\ttfamily]
class SpriteSquare extends NodeWithSize {
  
  Color color;
  Function handler;
  double left;
  double top;

  SpriteSquare(size, this.color, this.handler, this.top, this.left) : super(size){
    userInteractionEnabled = true;
  }

  @override handleEvent(SpriteBoxEvent event) {
    handler(event);
    return true;
  }

  @override
    Future<void> paint(Canvas canvas) async {
      canvas.drawRect(
        Rect.fromLTWH(left, top, size.height, size.width),
        Paint()..color = color
      );
    }
}
\end{lstlisting}

\textbf{handleAdd}
\begin{lstlisting}[language=Dart, basicstyle=\tiny\ttfamily]
handleAdd(SpriteBoxEvent event) {
    final _whitesquare = SpriteSquare(
      const Size(50.0, 50.0), 
      const Color(0xFFFFFFFF), 
      handleRemoveSelf, 
      Offset(event.pointerEvent.localPosition.dx, event.pointerEvent.localPosition.dy)
    );
    _background.addChild(_whitesquare);
    return true;
}
\end{lstlisting}
Can you help me understand why the square is not being drawn centered at the click location and suggest how to fix this issue?
}
\end{minipage}%
\hfill
\begin{minipage}{0.28\linewidth}
    \centering
    \vspace{0pt}
    \includegraphics[width=0.95\linewidth]{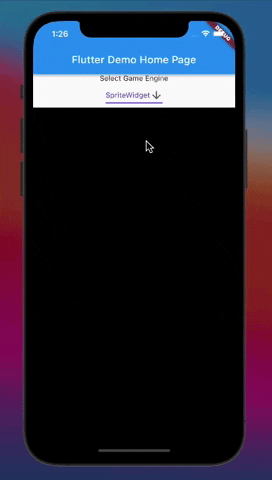}
\end{minipage}

\end{tcolorbox}

\begin{tcolorbox}[
  enhanced,
  sharp corners,
  colback=white,
  colframe=black,
  left=1pt, right=1pt, top=1pt, bottom=1pt,
  boxsep=0pt,
]
\scriptsize


\textcolor{blue}{\textbf{Response}}: 

{\tiny
The issue you're experiencing with the square not being drawn centered at the click location is likely due to the way you're calculating the position of the square. In your current implementation, you're using `Rect.fromLTWH` to draw the rectangle, which takes the left and top coordinates as the starting point of the rectangle. This means that the rectangle is drawn starting from the top-left corner at the specified coordinates, rather than being centered at those coordinates.

To center the square at the click location, you need to adjust the `left` and `top` values by subtracting half of the square's width and height, respectively. Here's how you can modify your code:

In the `handleAdd` function, calculate the `left` and `top` values like this:

\begin{lstlisting}[basicstyle=\tiny\ttfamily]
handleAdd(SpriteBoxEvent event) {
    final clickX = event.pointerEvent.localPosition.dx;
    final clickY = event.pointerEvent.localPosition.dy;
    
    // Calculate the top-left corner of the square so that it is centered at the click location
    final left = clickX - 25.0; // Half of the square's width (50.0 / 2)
    final top = clickY - 25.0;  // Half of the square's height (50.0 / 2)

    final _whitesquare = SpriteSquare(
      const Size(50.0, 50.0), 
      const Color(0xFFFFFFFF), 
      handleRemoveSelf, 
      left,
      top
    );
    _background.addChild(_whitesquare);
    return true;
}
\end{lstlisting}

By adjusting the `left` and `top` values in this way, the square will be centered at the click location. This is because you're effectively moving the top-left corner of the square to a position that centers the square around the click point.
}

\centering
\begin{tcolorbox}[
  enhanced,
  sharp corners,
  colback=gray!10,
  colframe=gray!10,
  boxrule=0pt,
  boxsep=1pt,
  left=0pt, right=0pt, top=0pt, bottom=0pt,
  width=\dimexpr\linewidth+1pt\relax 
]

\textbf{Category: Mobile \& Desktop——Dart}

\textbf{Evaluation Criteria: GPT-4o}

{\tiny
\textbf{Correctness Score: 35/50}\\
\textbf{Completeness Score: 25/50}
}

\textbf{Total Score: 60}

\end{tcolorbox}
\end{tcolorbox}

\caption{A case of InfiBench-V from the Mobile \& Desktop category. }
\label{fig:caseinfibench5}
\vspace{-4mm}
\end{figure}

\clearpage

\subsection{InfiBench-V Case 5}

\begin{figure}[h]
\centering
\scriptsize
\begin{tcolorbox}[
  enhanced,
  sharp corners,
  colback=white,
  colframe=black,
  left=1pt, right=1pt, top=1pt, bottom=1pt,
  boxsep=2pt,
  width=\linewidth
]
\scriptsize

\begin{minipage}{0.68\linewidth}
\textbf{Question:}  
{\tiny
In the context of a Ruby on Rails application using version 2.3.5 and Ruby 1.8.7, I am working with three database tables: A, B, and C. The relationships are as follows: table A has multiple associated records in table B, and table C also has multiple associated records in table B. There is no direct relationship between tables A and C.

I want to query records from table A and include associated records from tables B and C. I can include table B in the query using the following code:

\begin{lstlisting}[language=Ruby, basicstyle=\tiny\ttfamily]
A.all(:conditions => { :id => id }, :include => [:B])
\end{lstlisting}

However, I am unsure how to also include table C in the query. Despite extensive research, I have not found a solution. How can I modify the query to include records from both tables B and C when querying table A's records?
}
\end{minipage}%
\hfill
\begin{minipage}{0.28\linewidth}
    \centering
    \vspace{0pt}
    \includegraphics[width=0.95\linewidth]{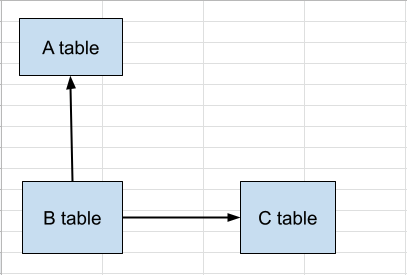}
\end{minipage}

\end{tcolorbox}

\begin{tcolorbox}[
  enhanced,
  sharp corners,
  colback=white,
  colframe=black,
  left=1pt, right=1pt, top=1pt, bottom=1pt,
  boxsep=0pt,
]
\scriptsize


\textcolor{blue}{\textbf{Response}}: 

{\tiny
To include associated records from both tables B and C when querying table A in a Ruby on Rails application, you need to ensure that the associations are set up correctly in your models. Assuming you have the following associations:

- Table A has many records in table B (e.g., \texttt{has\_many :bs}).\\
- Table C has many records in table B (e.g., \texttt{has\_many :cs}).

You can use eager loading with \texttt{:include} to load associated records from both tables B and C. Here's how you can modify your query:

\begin{lstlisting}[language=Ruby, basicstyle=\tiny\ttfamily]
A.all(:conditions => { :id => id }, :include => { :bs => :cs })
\end{lstlisting}

This assumes that:

- In model A, you have \texttt{has\_many :bs}. \\
- In model B, you have \texttt{belongs\_to :a} and \texttt{has\_many :cs}. \\
- In model C, you have \texttt{has\_many :bs}.

If the association between B and C is different, adjust the \texttt{:include} statement accordingly. If there is no direct relationship between A and C, but you want to include C through B, make sure the association is correctly defined in your models.

\textbf{Note:} Since you're using Rails 2.3.5 and Ruby 1.8.7, the syntax for eager loading is slightly different compared to newer versions of Rails. Make sure your models' associations are set up correctly to reflect these relationships.
}

\centering
\begin{tcolorbox}[
  enhanced,
  sharp corners,
  colback=gray!10,
  colframe=gray!10,
  boxrule=0pt,
  boxsep=1pt,
  left=0pt, right=0pt, top=0pt, bottom=0pt,
  width=\dimexpr\linewidth+1pt\relax 
]

\textbf{Category: Back-End——Ruby}

\textbf{Evaluation Criteria: GPT-4o}

{\tiny
\begin{enumerate}[left=0em, label={\arabic*.}, itemsep=0.3em, topsep=0pt, parsep=0pt]
    \item \textbf{Correctness (0–50):} The model answer provides a solution using Rails' eager loading feature with the \texttt{:include} option in a way that aligns with Rails 2.3.5 syntax. It correctly assumes that B acts as an intermediary between A and C. The proposed query modification \texttt{A.all(:conditions => \{ :id => id \}, :include => \{ :bs => :cs \})} is a valid way to include associated records from both tables B and C through B. This matches the associations depicted in the screenshot, where A is linked to B, and B is linked to C. The model answer elaborates on the associations needed in the models, which is technically correct. However, the reference answer is more concise and directly points to the usage of \texttt{:include => \{ :B => :C \}} without additional assumptions about model setup. The model answer's explanation about associations, while correct, is not necessary to resolve the specific query issue. Nonetheless, the model's proposed query would work given the correct associations. Therefore, the correctness of the model's answer is high, albeit slightly verbose.
    
    \item \textbf{Completeness (0–50):} The model answer provides a comprehensive response by explaining the underlying assumptions about the model associations, which are essential to achieving the desired query behavior. It goes beyond the reference answer by elaborating on how associations should be set up in the models. This additional detail provides a complete understanding of how the query interacts with the model structure. While the reference answer is more succinct, the model answer's thoroughness in explaining potential association configurations adds value and ensures that the user understands the prerequisites for the query to function as intended. Thus, the model answer is complete but includes information that, although helpful, is not strictly necessary to address the question directly.
\end{enumerate}

\vspace{0.3em}

\textbf{Scores:} \\
- Correctness: 45/50 \\
- Completeness: 45/50

}

\textbf{Total Score: 90}

\end{tcolorbox}
\end{tcolorbox}

\caption{A case of InfiBench-V from the Back-End category. }
\label{fig:caseinfibench6}
\vspace{-4mm}
\end{figure}



\newpage

\end{document}